\documentclass[pdflatex,sn-mathphys-num]{sn-jnl}

\usepackage{graphicx}%
\usepackage{multirow}%
\usepackage{amsmath,amssymb,amsfonts}%
\usepackage{amsthm}%
\usepackage{mathrsfs}%
\usepackage[title]{appendix}%
\usepackage{xcolor}%
\usepackage{textcomp}%
\usepackage{manyfoot}%
\usepackage{booktabs}%
\usepackage{algorithm}%
\usepackage{algorithmicx}%
\usepackage{algpseudocode}%
\usepackage{listings}%
\usepackage{pifont}
\usepackage{verbatim}
\usepackage{pifont}
\usepackage{multicol}
\usepackage{lscape}
\usepackage{float}      
\usepackage{placeins}   
\usepackage{capt-of}
\usepackage{graphicx}

\usepackage{subcaption}
\usepackage{hyperref}
\definecolor{gray1}{rgb}{0.95, 0.95, 0.96}

\theoremstyle{thmstyleone}%
\theoremstyle{thmstyletwo}%

\theoremstyle{thmstylethree}%

\begin{document}

\title{Span Modeling for Idiomaticity and Figurative Language Detection with Span Contrastive Loss}


\author[1]{\fnm{Blake} \sur{Matheny}}\email{matheny.blake@jaist.ac.jp}

\author[1]{\fnm{Phuong Minh} \sur{Nguyen}}\email{phuongnm@jaist.ac.jp}

\author[1]{\fnm{Minh Le} \sur{Nguyen}}\email{nguyenml@jaist.ac.jp}

\affil[1]{\orgdiv{Information Science}, \orgname{Japan Advanced Institute of Science and Technology}, \orgaddress{ \city{Ishikawa},  \country{Japan}}}

\abstract{ 
    The category of figurative language contains many varieties, some of which are non-compositional in nature. This type of phrase or multi-word expression (MWE) includes idioms, which represent a single meaning that does not consist of the sum of its words. For language models, this presents a unique problem due to tokenization and adjacent contextual embeddings. Many large language models have overcome this issue with large phrase vocabulary, though immediate recognition frequently fails without one- or few-shot prompting or instruction finetuning. The best results have been achieved with BERT-based or LSTM finetuning approaches. The model in this paper contains one such variety. We propose BERT- and RoBERTa-based models finetuned with a combination of slot loss and span contrastive loss (SCL) with hard negative reweighting to improve idiomaticity detection, attaining state of the art sequence accuracy performance on existing datasets. Comparative ablation studies show the effectiveness of SCL and its generalizability. The geometric mean of F1 and sequence accuracy (SA) is also proposed to assess a model's span awareness and general performance together. 
}




\maketitle

\section{Introduction}\label{sec1}

Idiomaticity detection remains a crucial task for language models in natural language understanding (NLU). Figurative language, like idioms, frequently evades detection by even the largest models. Idioms comprise a subset of figurative language representing a single semantic unit in multiple words, which can be compositional or non-compositional. Unless a language model contains these as multi-word expressions explicitly in their pretraining and vocabulary, even a very large language model may fail the detection task. Our work builds upon prior research and datasets developed for the task of idiomaticity and figurative language detection \cite{matheny2025nlpdatasetsidiomfigurative}. 
Concerted efforts using contrastive loss have focused on similarity objectives with a margin as a tunable hyperparameter \cite{he-etal-2024-enhancing}. This line of research demonstrated improvements over existing methods in terms of F1 score, although the benefits of hard negative mining remained limited. With the introduction of margin adjustment in the similarity-based loss, performance further improved. Training focused on anchor-negative and positive-negative pairs, equating anchors and positives. This effectively doubled the number of training samples by swapping positives and anchors. Subsequent testing evaluated model predictions against human similarity judgments using Semantic Textual Similarity (STS) data from SemEval2022 \cite{tayyarmadabushi2022semeval2}. While our approach follows a similar contrastive learning structure, we focus on span representation rather than sentence representations and pair-wise comparisons.

To address these challenges in slot labeling, span contrastive loss (SCL) with weighted hard negatives was developed and combined with slot loss to train BERT- and RoBERTa-based models on both new and widely used idiom and figurative language datasets, including MAGPIE-idioms \cite{haagsma-etal-2020-magpie} and \textit{P}IFL-OSCAR \cite{matheny2025nlpdatasetsidiomfigurative}. This research focused not only on objective functions and loss combinations, but also on similarity measurements of target span representations, computed as the average of word embeddings within each span.
Our contributions are: 
\begin{enumerate}
    \item Span contrastive loss employed for multi-word logic
    \item Cross evaluation on multi-word expresssion (MWE) and figurative language datasets to examine generalizability on diverse datasets seen quantitatively in tables and qualitatively in figures 
    \item Proposing geometric mean of F1 and sequence accuracy as span-aware performance
    \item State of the art performance for sequence accuracy on the MAGPIE dataset (evaluation under \textit{random} split setting)
\end{enumerate}


\section{Related Work}
This work represents a combination of techniques, building on adaptations of existing architectures and a novel application of joint loss. As such, the relevant literature can be organized into four principal components:
(1) Contextual sequence encoders for span and slot representations, (2) Contrastive representation learning and supervised contrastive objectives,  (3) Hard negatives and adaptive negative weighting strategies, and (4) Figurative language and idiom understanding (task context).

\paragraph{Contextual sequence encoders for span and slot representations}
Modern natural language understanding (NLU) systems rely heavily on contextual sequence encoders, with Transformer architectures providing a general-purpose mechanism for producing context-sensitive token representations \cite{vaswani2017attention}. BERT-style pretraining further improves slot labeling by supplying bidirectional contextual features that can be fine-tuned for task-specific objectives \cite{devlin2019bert}.
For structured sequence labeling, globally normalized decoding with conditional random fields (CRFs) remains a standard technique for enforcing label consistency, particularly in NER-like span prediction settings \cite{lample-etal-2016-neural, lafferty2001crf}. In parallel, phrase- and span-centric encoders such as PhraseBERT demonstrate that explicitly modeling multi-token phrases yields improved semantic representations for span-level tasks, suggesting that phrase-aware objectives can enhance token-level encoders \cite{Wang2021PhraseBERT}. Our approach aligns with this direction by deriving span representations from contextual token embeddings and directly supervising them through both classification and contrastive losses.

\paragraph{Contrastive representation learning and supervised contrastive objectives}
Contrastive learning has emerged as a powerful framework for shaping representation geometry by pulling semantically related instances together and pushing unrelated instances apart, often using temperature-scaled softmax objectives such as InfoNCE \cite{oord2019representationlearningcontrastivepredictive}. Large-batch and in-batch negative formulations have shown that contrastive objectives can scale effectively while producing robust representations \cite{chen2020simpleframeworkcontrastivelearning}.
In supervised settings, contrastive learning (SupCon) generalizes this framework by defining positive pairs with shared labels and negatives with differing labels, encouraging class-conditional clustering that benefits classification \cite{khoslaNEURIPS2020_d89a66c7}. The span-level contrastive component of our method follows this supervised formulation: span embeddings derived from contextual encoders are contrasted using gold slot labels, complementing the standard token-level cross-entropy (and optional CRF) objectives.

\paragraph{Hard negatives and adaptive negative weighting strategies}
A consistent finding across metric and contrastive learning is that negative sampling strongly influences optimization. Metric learning objectives such as the N-pair loss \cite{sohn2016npair} and triplet loss variants highlight the importance of informative negatives. More recent work, mentioned previously, proposes adaptive contrastive triplet losses, which dynamically adjust margins or weights based on similarity or difficulty, improving convergence while reducing sensitivity to poorly chosen negatives \cite{he-etal-2024-enhancing}.
Analyses of contrastive optimization further show that hard negatives--those close to the anchor in representation space--provide stronger learning signals but may introduce instability or false negatives if used indiscriminately \cite{robinson2020hardneg-arxiv, robinson2021hardneg}. As a result, several methods emphasize adaptive weighting or curriculum-style treatment of negatives rather than binary selection \cite{NEURIPS2020_f7cade80kalantadis}. Our approach follows this direction by identifying confusable negative spans via in-batch similarity and increasing their contribution through reweighting, preserving global separation signals while focusing gradient pressure on close cases.

\paragraph{Figurative language and idiom understanding}
Although our modeling choices are general-purpose, they connect naturally to prior work on figurative language understanding, where surface overlap and contextual ambiguity are common. Recent studies on idiom identification and non-compositional phrase interpretation demonstrate both the strengths and limitations of pretrained contextual encoders, motivating the need for additional inductive bias or training signals beyond standard fine-tuning \cite{chakrabarty-etal-2022-flute, yayavaram-etal-2024-bert}.
Metaphor detection and figurative language benchmarks consistently demonstrate that contextualized language representations substantially improve performance over static or feature-based approaches, particularly in capturing semantic dependencies across sentence contexts \cite{hall-maudslay-etal-2020-metaphor, liu-etal-2020-metaphor}. However, prior work also highlights that figurative language tasks remain challenging due to the high degree of semantic overlap between figurative and literal usages, leading to confusable representations under standard training objectives \cite{wang-etal-2023-metaphor}. These findings suggest that while contextual encoders provide a strong foundation, they benefit from task-specific learning signals that promote semantic separability more directly. From this perspective, span-level supervised contrastive learning with adaptive hard-negative emphasis for slot labeling offers a general mechanism for improving representational discrimination in precisely the kinds of ambiguous and non-literal instances that symbolize figurative language understanding.

\section{Datasets}

A variety of datasets of diverse sizes and classifications have been used in idiomaticity detection research. This research examined a selection of the most recent, with labeling adaptations applied where necessary for experimentation \cite{matheny2025nlpdatasetsidiomfigurative}. 
The list includes the human annotated datasets: IFL-C4-A (derived from C4 and Common Crawl) \cite{matheny2025nlpdatasetsidiomfigurative}, IFL-OSCAR-A (derived from OSCAR and Common Crawl) \cite{matheny2025nlpdatasetsidiomfigurative}, MAGPIE (Crowded sourced, BNC derived) \cite{haagsma-etal-2020-magpie}, VNC (BNC derived) \cite{VNC-cook-fazly-stevenson-2008-vnctokens}, EPIE (BNC derived) \cite{saxena2020epie}, SemEval 2022 Task 2 (news and literature sources) \cite{tayyarmadabushi2022semeval2}, and FLUTE (stories, news) \cite{chakrabarty-etal-2022-flute}. One large dataset of intentionally \textit{potential} idioms and figurative language (\textit{P}IFL-OSCAR) was included due to highly generalizable baseline training \cite{matheny2025nlpdatasetsidiomfigurative}.  EPIE and VNC were not used in this research. Additional details for each dataset can be seen in Table~\ref{tab:current_sota}.

A number of state-of-the-art results in Table \ref{tab:current_sota} confirm persistent research endeavors into this field of abstract sentiment and figurative language. Sequence accuracy, token accuracy, and macro F1 represent the most common metrics for these training scenarios.  While this research contained experiments pertaining to each of these datasets, the main focus will be upon MAGPIE (random) and \textit{P}IFL-OSCAR. Our previous research outperformed Yayavaram's sequence accuracy results \cite{yayavaram-etal-2024-bert}. The results in this paper outperformed all previous research on sequence accuracy on MAGPIE (random).\footnote{Further details regarding the datasets can be found in previous research. \cite{matheny2025nlpdatasetsidiomfigurative}}

\begin{table*}[htbp] 
\caption{Current SOTA of idiom and figurative language datasets. The asterisk (*) denotes that the result is taken from the corresponding dataset paper.}
\resizebox{\linewidth}{!}{%

\begin{tabular}{l l l l }
\toprule
\textbf{Dataset} & \textbf{Task} & \textbf{Primary} & \textbf{Best} \\
\midrule
EPIE (Formal) & Span tagging (BIO) & Acc. & \textbf{98.00} *\\
MAGPIE (random)& Span id. (exact) & Sequence Accuracy& \textbf{91.51} \cite{yayavaram-etal-2024-bert}\\
 MAGPIE (random)& Span id.& Sequence Accuracy&92.97 \cite{matheny2025nlpdatasetsidiomfigurative}\\
MAGPIE (type-aware) & Span id. (unseen types) & Sequence Accuracy& \textbf{70.47} \cite{disc-zeng2021idiomatic}\\
\midrule
 SemEval-2022 2A & Idiomaticity existence& Macro-F1 (All)& \textbf{93.85} *\\
 FLUTE (Idioms)& Sentence-pair entailment& Accuracy& \textbf{79.20} *\\\midrule
 IFL-C4-A& Slot labeling& F1&96.91 \cite{matheny2025nlpdatasetsidiomfigurative}\\
 
 \textit{P}IFL-OSCAR& Slot labeling& F1&98.55 \cite{matheny2025nlpdatasetsidiomfigurative}\\\bottomrule
\end{tabular}
}
\label{tab:current_sota}
\end{table*}

\section{Methods}

Our modeling framework followed the joint intent–slot formulation introduced by \citet{chen2019bert-intent-slot} and further adapted in JointBERT CAE \cite{phuong2022cae}. Similar to these prior approaches, contextual token representations were obtained via Eq.~\ref{eq:context_tok}, and token-level predictions were generated through Eq.~\ref{eq:token_lb}.  Building on our previous span-aware modeling efforts, this framework was extended by incorporating a span contrastive component (Eq.~\ref{eq:spand_contrastive_loss}) with the base objective remaining the averaged cross-entropy loss (Eq.~\ref{eq:slot_loss}), which explicitly encourages separation between semantically ambiguous span representations while preserving the underlying sequence labeling formulation established in earlier work. 

\begin{align}
    \mathbf{h}^{[CLS]}, \mathbf{h}_i^{\text{word}} &= 
        \mathrm{BERT}(s) \label{eq:context_tok} \\
    \hat{\mathbf{y}}_i &= 
        \mathrm{softmax}\!\left(
        \mathbf{W}\,\mathbf{h}_i^{\text{word}} + \mathbf{b}
        \right)  \label{eq:token_lb}  \\
    \mathcal{L}_{\text{slot}} &=
        \frac{1}{|s|}
        \sum_{i=1}^{|s|}
        \mathrm{CrossEntropy}
        \big(
        \hat{\mathbf{y}}_i,
        \mathbf{y}_i
        \big) \label{eq:slot_loss} \\
    \mathcal{L}_{\text{total}} &=
        \mathcal{L}_{\text{slot}}
        +
        \lambda_{\text{span}} \times \mathcal{L}_{\text{span contrastive}} \label{eq:spand_contrastive_loss}
\end{align}
where ${s}$ is the input sentence, $\mathbf{W}, \mathbf{b}$ is learnable parameters, $\lambda_{span}$ is the controllable hyper-parameter. 

\subsection{Span Contrastive Loss}
Adapting this architecture, we also further improved it with the span contrastive objective. Span contrastive architecture has proven to enhance the subtle variations between near negatives and positives in embedding space. The formalizations below express the nuance between regular span contrastive loss and our reweighted span contrastive loss for hard negatives \textit{(SCL-neg)}. To model the spans, we represented them as the mean of the token embeddings contained in the labeled or masked span.
\paragraph{Regular Span Contrastive Loss}

Regular span contrastive learning and our hard-negative reweighted variant were differentiated with a figurative example with idioms. Three span types exist:

\begin{itemize}
    \item \textsc{Idiom/Figurative MWE}: non-compositional, figurative meaning
    \item \textsc{Hard negative}: literal, compositional usage of the same surface form
    \item \textsc{Easy negative}: spans with dissimilar surface semantics and labels
\end{itemize}

A supervised span-level contrastive objective was adopted, where each anchor span
is contrasted against all other spans in the minibatch.
Let \(\ell_{ij}\) denote the similarity logit between span \(s_i\) and span \(s_j\),
scaled by a temperature \(\tau\). 
Let \(\mathbf{z}_i, \mathbf{z}_j\)  denote the span vector representation of  \(s_i, s_j\), respectively, by average representations of words computed by the language model:

\[
\ell_{ij} = \frac{\mathrm{sim}(s_i, s_j)}{\tau} = \frac{\mathbf{z}_i^\top \mathbf{z}_j} {\tau}.
\]

For each anchor \(s_i\), let \(P(i)\) denote the set of positive spans
(those sharing the same label), and let all remaining spans be treated as negatives.
The baseline supervised contrastive loss is:

\[
\mathcal{L}^{\text{reg}}_i
=
-\log
\frac{
\sum_{p \in P(i)} \exp(\ell_{ip})
}{
\sum_{j \ne i} \exp(\ell_{ij})
}.
\]

\vspace{0.5em}
\paragraph{Span contrastive loss with hard-negative reweighting}
To strengthen the contribution of difficult negatives, the top-\(k\)
hardest negative spans for each anchor were selected.  \(\text{TopKNeg}(i)\) denotes the set of up to \(k\) negative spans with the
highest similarity to \(s_i\) (e.g., $k = 5$).
Rather than modifying the numerator, we amplify these hard negatives in the denominator:

\[
\mathcal{L}^{\text{hard}}_i
=
-\log
\frac{
\sum_{p \in P(i)} \exp(\ell_{ip})
}{
\sum_{j \ne i} \exp(\ell_{ij})
+
\sum_{j \in \text{TopKNeg}(i)} \exp(\ell_{ij}).
}.
\]

\paragraph{Final objective.}
In our implementation, the final span contrastive loss is the average of the
baseline and hard-negative losses:

\[
\mathcal{L}_i
=
\frac{1}{2}\mathcal{L}^{\text{reg}}_i
+
\frac{1}{2}\mathcal{L}^{\text{hard}}_i.
\]

The overall span contrastive loss ($\mathcal{L}_{\text{span contrastive}}$) is computed by averaging over all anchor spans
that have at least one positive counterpart in the minibatch.

\paragraph{Concrete example of span contrastive losses}
For a clearer understanding of the difference between regular span contrastive loss and hard-negative contrastive loss, a concrete example was presented along with the corresponding formulations of these losses.
Consider the sentence: \emph{``The skeptic finally \underline{saw the light}.''} Suppose that span extraction within a minibatch yields:

\begin{itemize}
    \item $s_1$: \emph{saw the light} (idiomatic usage, label \textsc{IDIOM})
    \item $s_4$: \emph{saw the light} (literal usage in another sentence, label \textsc{Hard negative})
    \item $s_5$: \emph{jump through hoops} (idiomatic paraphrase, label \textsc{IDIOM})
    \item $s_2, s_3, s_6$: unrelated spans (label \textsc{Easy negative})
\end{itemize}

The regular contrastive loss for an anchor {$s_{i=1}$} is computed as follow:
\[
\mathcal{L}^{\text{reg}}_1
=
-\log
\frac{\exp(\ell_{1,5})}
{\exp(\ell_{1,5})
+ \exp(\ell_{1,4})
+ \exp(\ell_{1,6})
+ \exp(\ell_{1,2})
+ \exp(\ell_{1,3})}.
\]

The new span contrastive loss emphasizes the separation of the hard negative by reweighting it in the denominator. 
Although $s_4$ (\textsc{Hard negative}) is the most confusable negative, it is treated identically to unrelated spans. As a result, the loss provides only a weak signal to separate literal and idiomatic realizations of the same surface form, allowing their representations to remain close in embedding space.
This yields the hard-negative contrastive loss:

\[
\mathcal{L}^{\text{hard}}_1
=
-\log
\frac{\exp(\ell_{1,5})}
{\exp(\ell_{1,5})
+ 2\exp(\ell_{1,4})
+ 2\exp(\ell_{1,6})
+ 2\exp(\ell_{1,2})
+ 2\exp(\ell_{1,3})}.
\]
\vspace{0.5em}
 
\section{Evaluation Metrics}
Common metrics, as those indicated in Table \ref{tab:current_sota} of current SOTA, are used to measure model performance. Using true positives, false positives, and false negatives, precision and recall allow researchers to determine a model's accuracy. Combining these into a composite score, $F1$, further shows the general ability the model. As idioms and figurative language typically contain multiple words, sequence accuracy also maintains importance.
\paragraph{Sentence-level Accuracy} 
For model performance on slot labeling spans accurately, we employ sentence-level accuracy (also referred to as sequence accuracy). It is defined by the average of correctly labeled sequences, where $\hat{y}_i$ denotes the prediction for the $i$-th sample in the evaluation dataset containing a total of $N$ samples:

\newcommand{\I}{\mathbf{1}}

\begin{align}
\mathrm{Sequence\, Accuracy\, (SA)}
= \frac{1}{N}\sum_{i=1}^{N} \I\{\hat{y}_i = {y}_i\}
\end{align}
\paragraph{Entity-level performance}  
Given $\hat{E}_i, E_i$ which denote the set of predicted entities and gold entities for the $i^{th}$ sample in the evaluation dataset, the standard metrics to assess LLM performance include F1, precision, and recall.  

\newcommand{\Egoldi}{E_i}
\newcommand{\Epredi}{\hat{E}_i}

\newcommand{\TP}{\mathrm{TP}}
\newcommand{\FP}{\mathrm{FP}}
\newcommand{\FN}{\mathrm{FN}}

\begin{align}
\TP &= \sum_{i=1}^{N} \bigl|\, \Egoldi \cap \Epredi \,\bigr|, \\
\FP &= \sum_{i=1}^{N} \bigl|\, \Epredi \setminus \Egoldi \,\bigr|, \\
\FN &= \sum_{i=1}^{N} \bigl|\, \Egoldi \setminus \Epredi \,\bigr|.
\end{align}
All positive results are examined when using precision, {P}. The relevant and retrieved instances are measured using \textbf{TP}, true positives and \textbf{FP}, false positives as follows:
\begin{align}
\mathrm{Precision}\ (P) &= \frac{\TP}{\TP + \FP}. 
\end{align}

False positives take the place of false negatives, thereby allowing recall. {R}, to focus attention on relevance by:
\begin{align}
    \mathrm{Recall}\ (R)    &= \frac{\TP}{\TP + \FN}. 
\end{align}
Recall and precision measure relevance and retrieval and their relationship to positives and negatives, but F1 can show the general ability of a model.
 For precision, \textbf{P}, and recall, \textbf{R}:
\begin{align}
    \mathrm{F1}             &= \frac{2PR}{P + R}.
\end{align}
\paragraph{Refined Span Metrics\label{sec:refined_span_metrics}}
The previous metrics are reliable for traditional evaluation techniques, relating positive and negative results with a logical composite score of precision and recall. For our span-oriented task, we supply optional metrics to clearly incorporate generalizability and span-awareness. As such, the understanding of each model's performance across datasets of varying sizes and distributions was augmented.  The average performance of a metric $m$, $\mu_m$ was also introduced, where both span accuracy (SA) and F1 score are applicable for $K$ datasets:
\begin{align}
\mu_m &= \mathbb{E}[m] \quad \text{such that} \quad  m \in \{ \text{SA}, \text{F1} \}  \\
\mu &= \frac{1}{K} \sum_{i=1}^{K} F1_i,    
\end{align}

Further, $R_m$ was defined as the worst performance on metric $m$. Consistent with the use of span accuracy and F1, we define
\[
R_m = \min(m).
\]

As this research emphasized the importance of span both internally and externally to the model architecture, the use of the geometric mean of F1 score (general performance) and sequence accuracy (SA) as a measure of span awareness and effectiveness was proposed. This is defined as
\[
\mathrm{GM} = \sqrt{\mathrm{F1} \cdot \mathrm{SA}}.
\]
\section{Experiments}

Experiments were performed to test the efficacy of the combined slot loss and labeling architecture with span contrastive loss and negative reweighting. Baseline evaluations and cross evaluation results of only slot loss are contained within the paper introducing \textit{P}IFL-OSCAR \cite{matheny2025nlpdatasetsidiomfigurative}. This research focused on the aforementioned span contrastive loss with negative reweighting. Hence, the experiments performed include the coefficient of SCL ablation and cross evaluation of the highest performing coefficient.  As seen above, slot loss and span contrastive loss were combined with the latter having a variable coefficient. Training and evaluations were conducted on each rational number from 0.0 to 1.0, by 0.1 intervals. Subsequently, the best performing models were chosen and labeled by the correlating SCL coefficient. Then, comparisons between baseline and ablated results, training efficiency, and metric introduction were also shown.

All experiments were conducted on a 40GB A100 GPU. Maximum accuracy was achieved at a batch size of 4. For each model training on a smaller dataset, the logging and evaluation steps were set to 10, while for larger datasets, 1000 was necessary. Each model was cross evaluated on all datasets under two training settings:  (1) the baseline setting, which did not use span contrastive loss ($\lambda_{span} = 0$), denoted as \texttt{SCL\_neg\_0.0}; and (2) a setting using span contrastive loss with a tuned hyperparameter $\lambda_{span}$, denoted as \texttt{SCL\_neg\_}{$\lambda_{span}$}.

\subsection{Results and Analysis}
\begin{figure}[!htbp]
    \centering
    \begin{minipage}{0.33\textwidth}
        \centering
    \includegraphics[width=\linewidth]{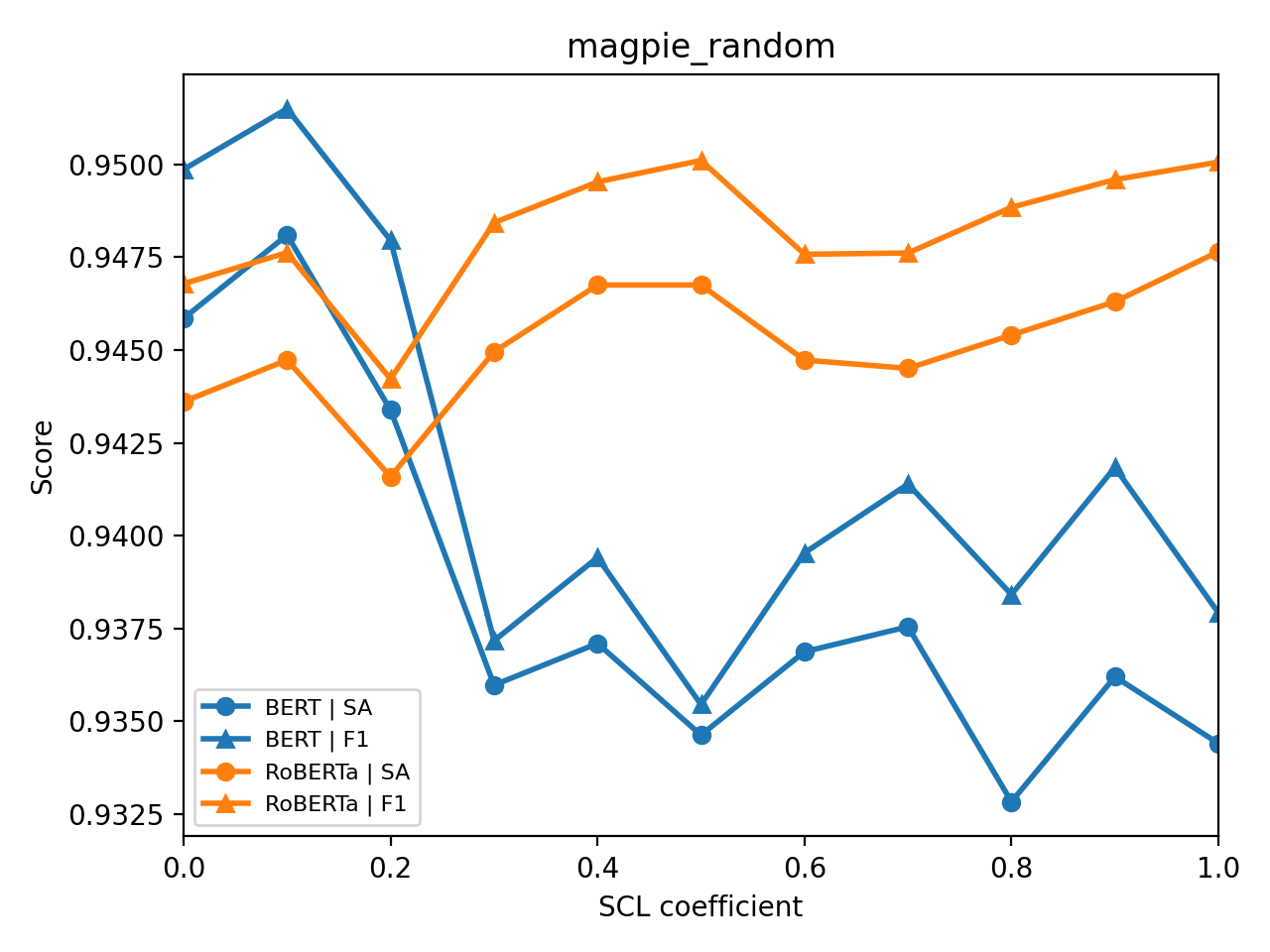}

\end{minipage}\hfill
    \begin{minipage}{0.33\textwidth}
            \centering
        \centering
    \includegraphics[width=\linewidth]{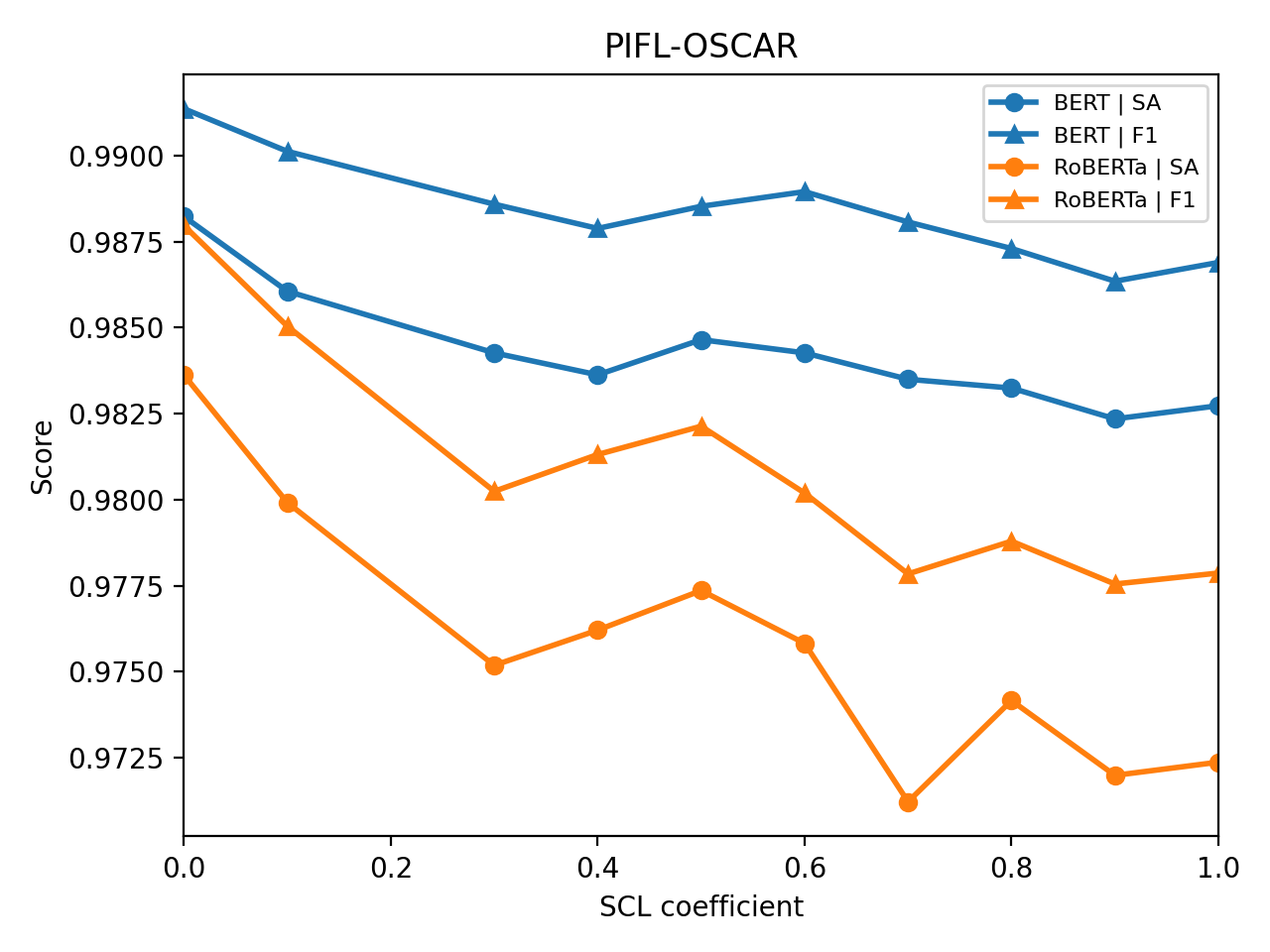}
    
\end{minipage}\hfill
    \begin{minipage}{0.33\textwidth}
            \centering
        \centering
    \includegraphics[width=\linewidth]{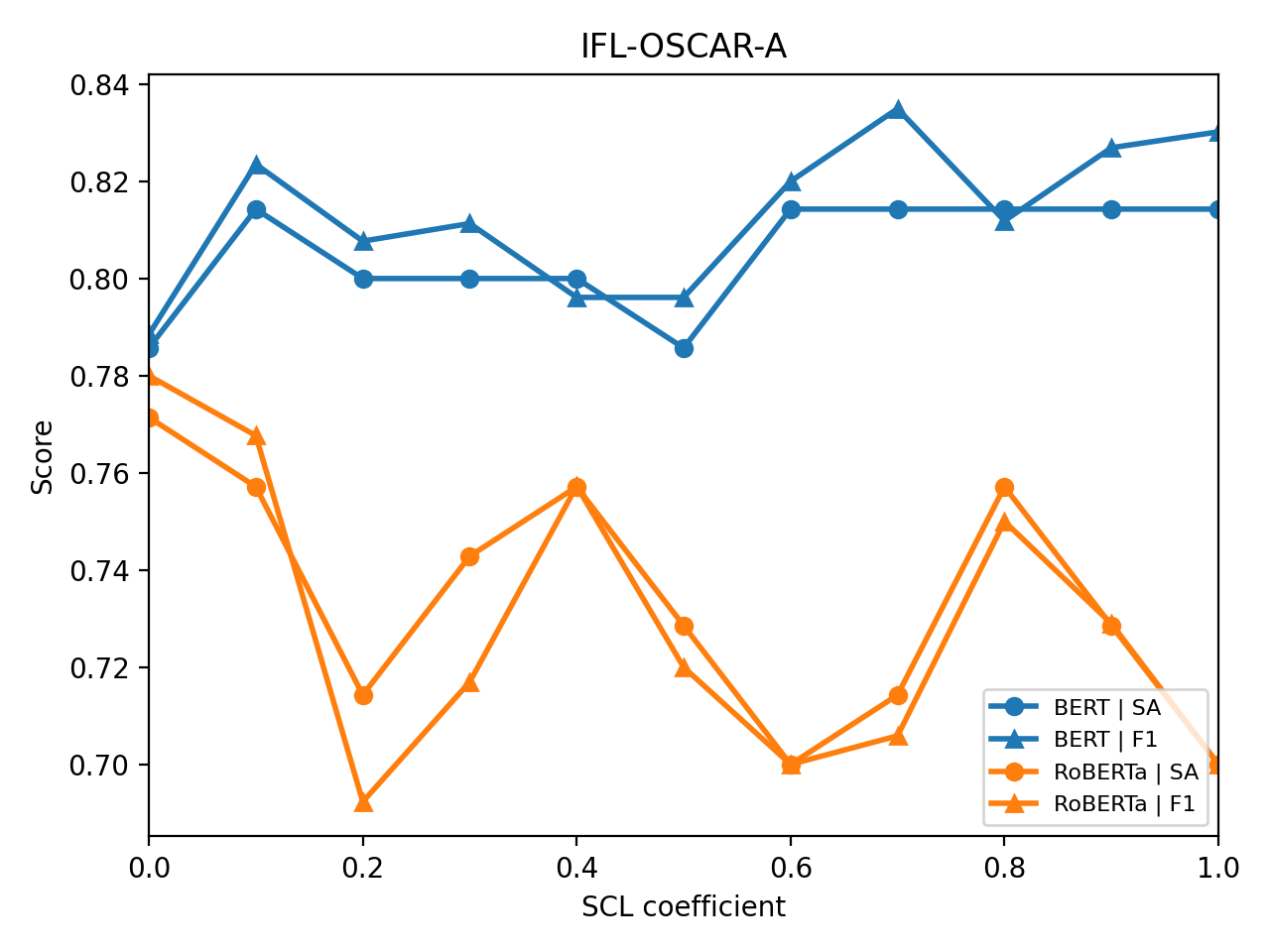}
    
\end{minipage}\hfill
    \begin{minipage}{0.33\textwidth}
            \centering
        \centering
    \includegraphics[width=\linewidth]{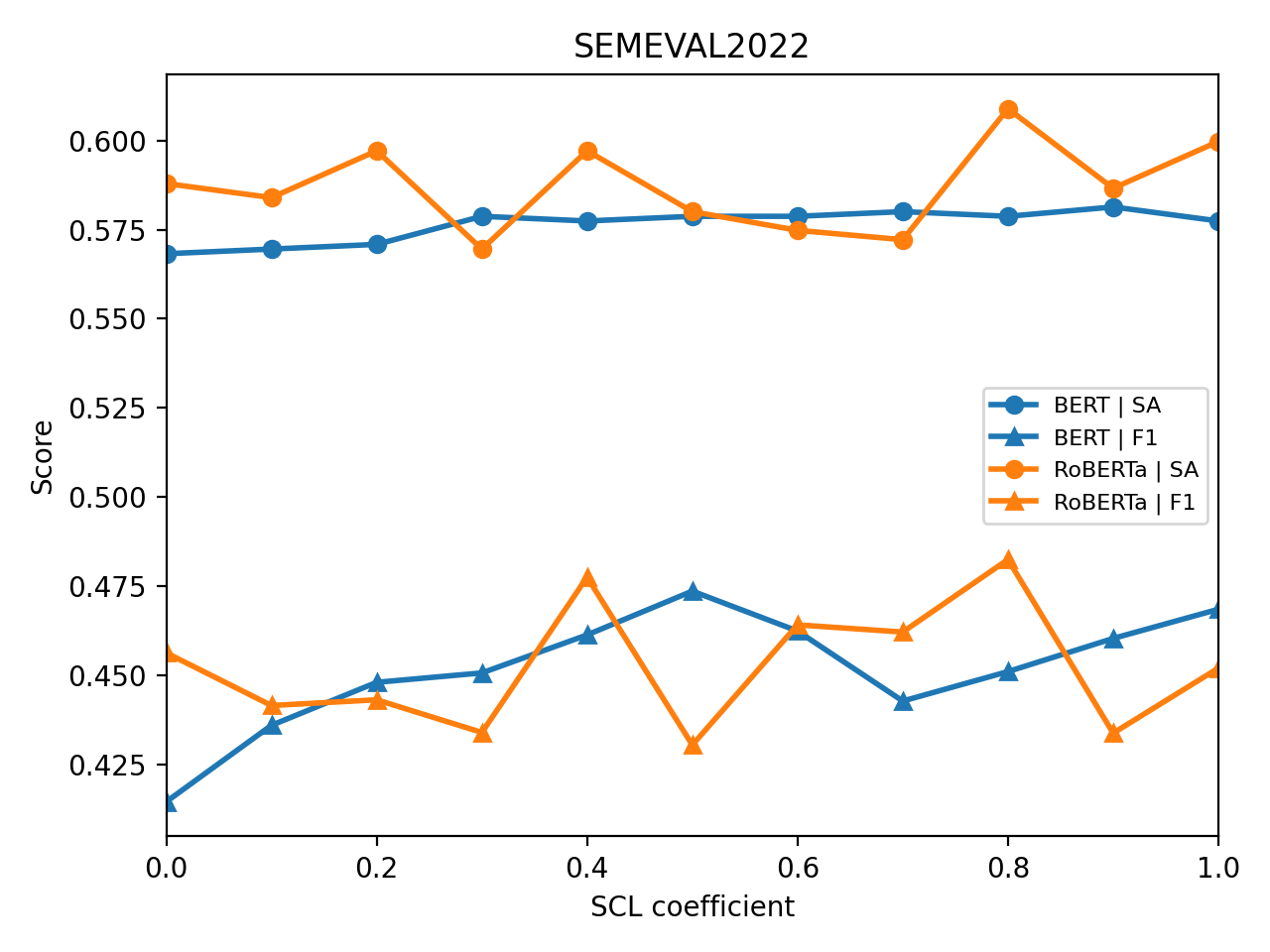}
    
\end{minipage}\hfill
    \begin{minipage}{0.33\textwidth}
            \centering
        \centering
    \includegraphics[width=\linewidth]{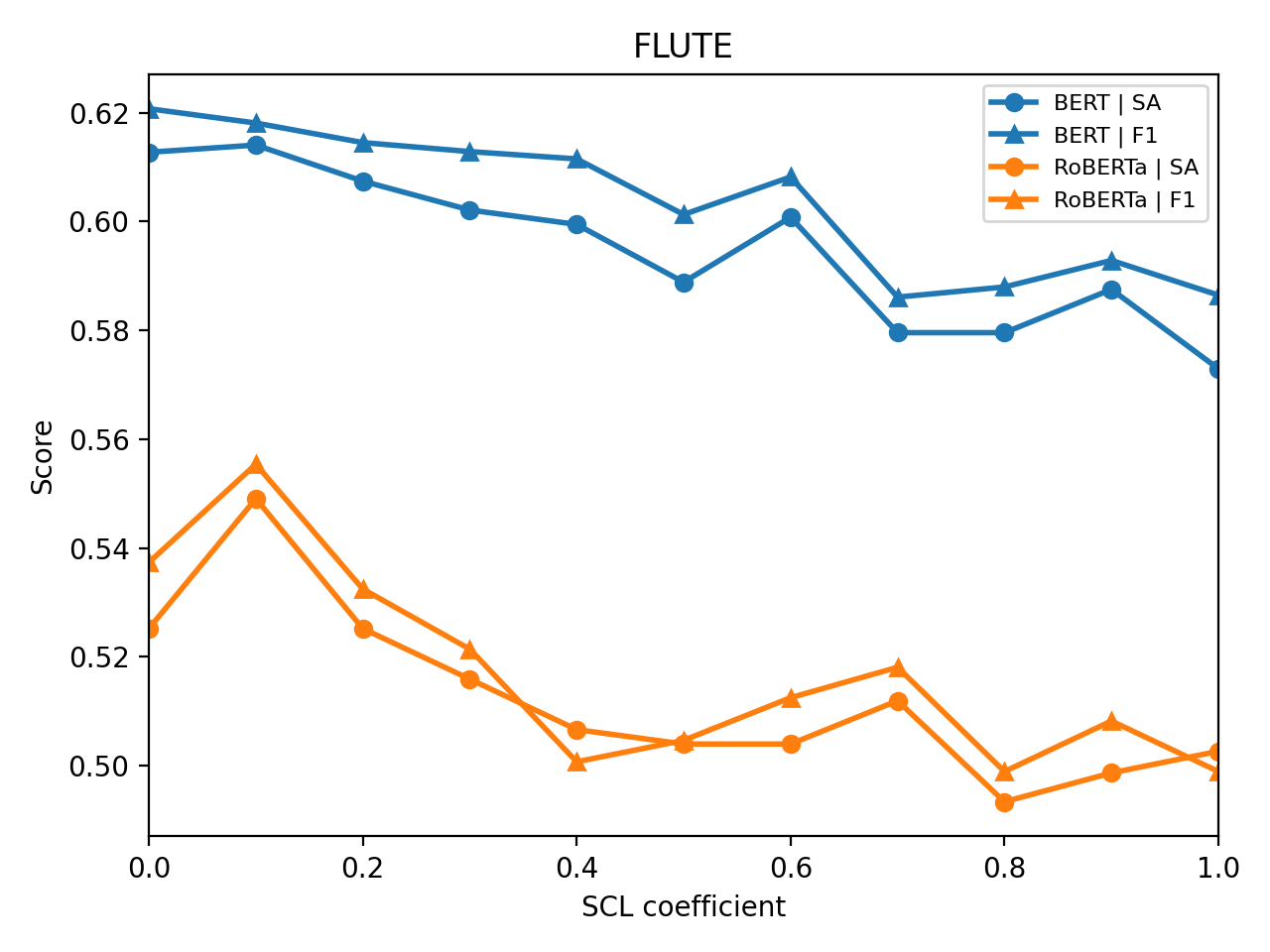}
    
\end{minipage}\hfill
    \begin{minipage}{0.33\textwidth}
            \centering
        \centering
    \includegraphics[width=\linewidth]{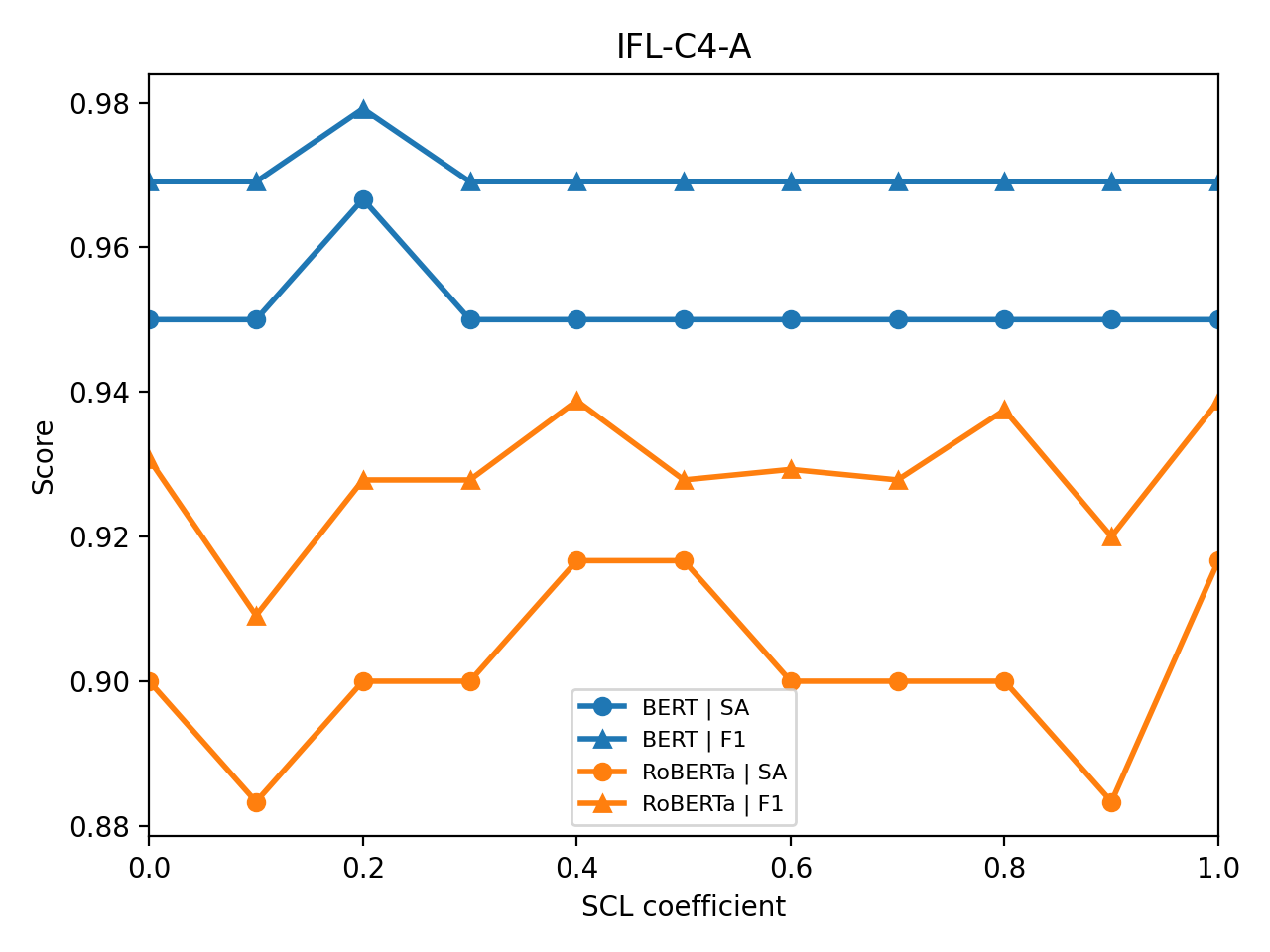}

  \end{minipage}
  \caption{Coefficient ablation results for all datasets. \label{fig:ablation_grid}}
\end{figure}

Experiments on this slot loss architecture of contrastive loss with hard negatives used both BERT and RoBERTa. These were finetuned on each dataset with a training/development/test split of 80/10/10 percent, respectively.
Training optimized them for slot precision on the development set. Evaluation occurred on the remaining 10 percent test set. An ablation study was also performed to determine the best coefficient for sequence accuracy. The study showed optimal performance over various coefficients by model and dataset. Ablation results for F1 score and sequence accuracy can be seen in Fig. \ref{fig:ablation_grid}.  

\begin{figure}[!htbp]
    \centering
    \includegraphics[width=1\linewidth]{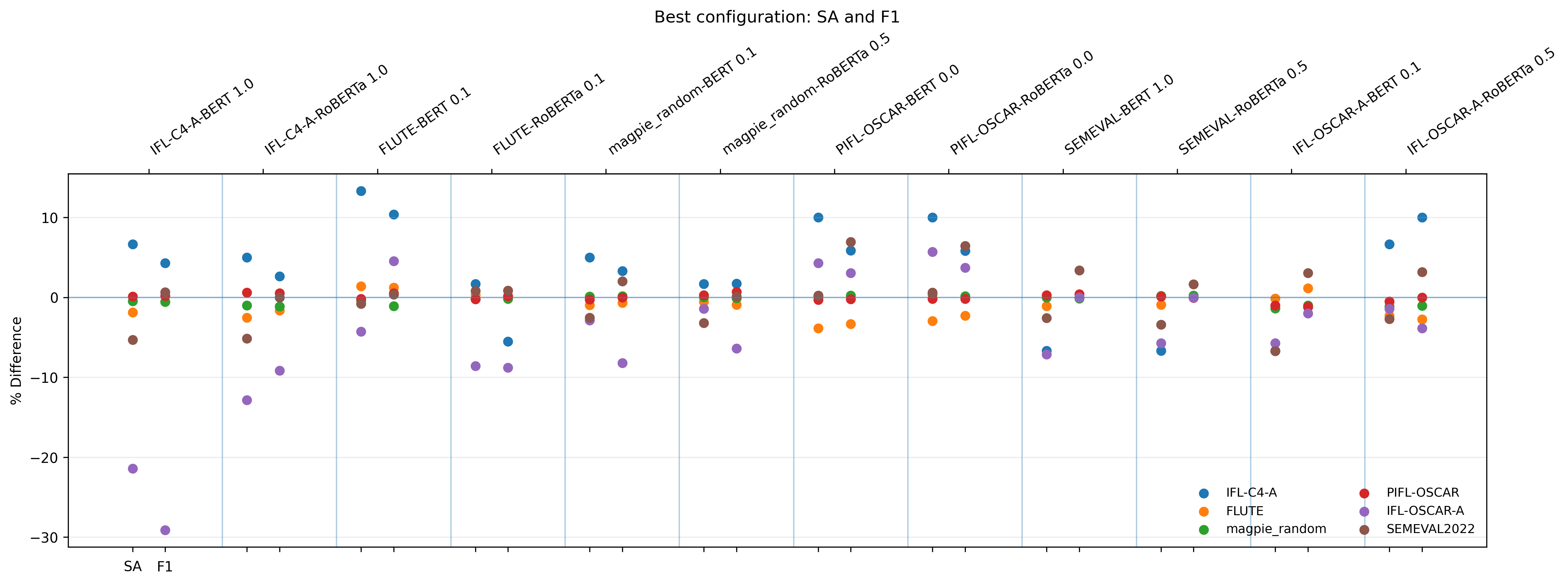}
    \caption{Difference between test and development set sequence accuracy and F1 evaluation for best coefficient of each model with \textit{SCL-neg architecture}.}
    \label{fig:best_stab}
\end{figure}

\paragraph{Generalization from Development to Test} 
One of the first positive training indications for any model showed the test set results improved over the development set results ($m_{test} - m_{dev}\, \textrm{where}\, m \in \{F1, SA\}$). The best settings for span contrastive loss coefficient are seen in Fig. \ref{fig:best_stab}, showing training stability for each. Evaluations on the test set for each model were consistently higher, while BERT-trained models, rather than RoBERTa-trained models performed better for all metrics, as seen in Fig.  \ref{fig:best_stab}. The results indicate a highly stable training architecture showing better performance for all metrics on the test set.

\begin{figure}[!htbp]
    \centering
    \includegraphics[width=0.94\linewidth]{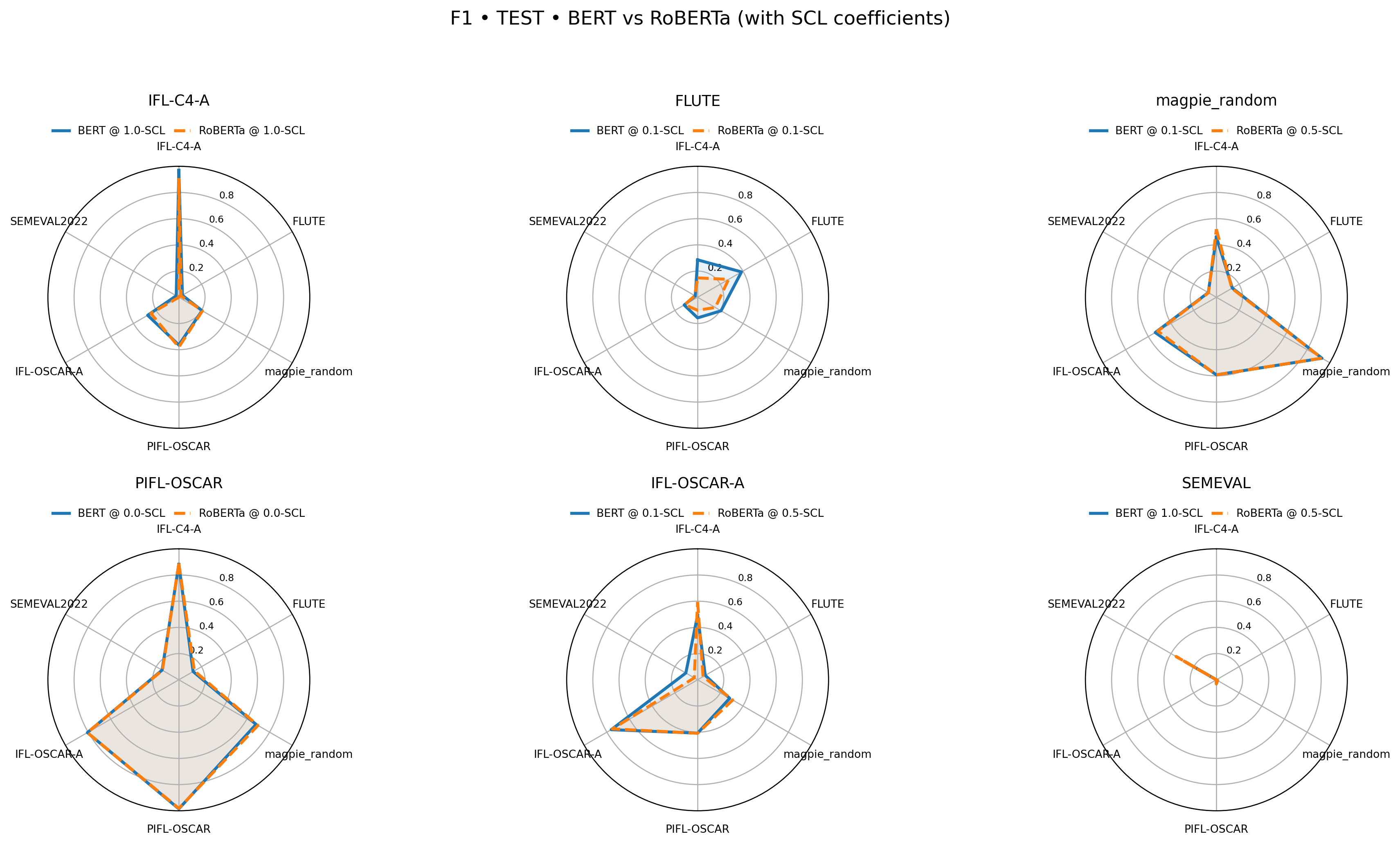}
    \caption{Visualization of F1 cross evaluation results on the test sets of each dataset.}
    \label{fig:test_f1_best}
\end{figure}

\begin{figure}[!htbp]
    \centering
    \includegraphics[width=0.94\linewidth]{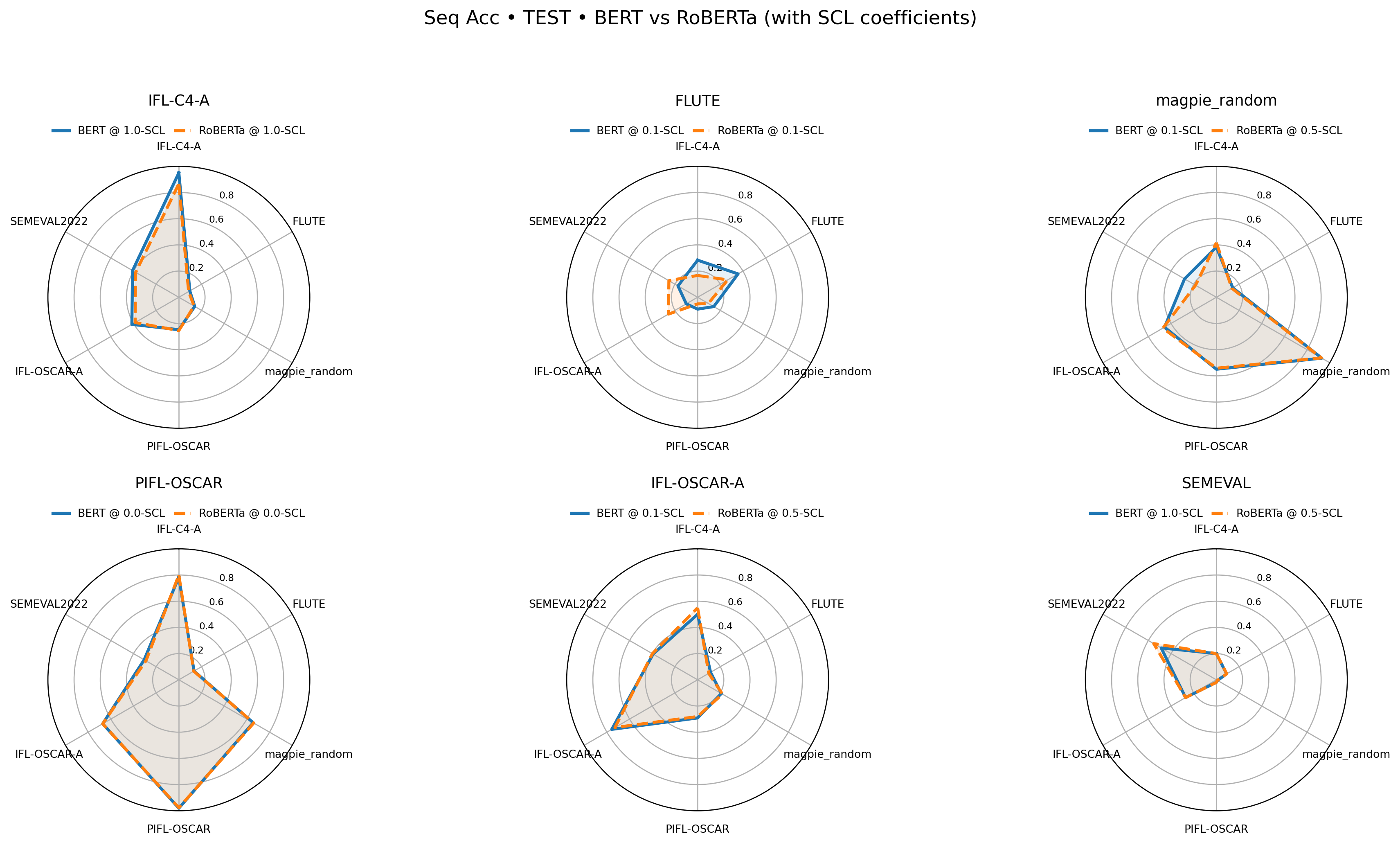}
    \caption{Visualization of SA cross evaluation results on the test sets of each dataset.}
    \label{fig:test_sa_best}
\end{figure}

\paragraph{Generalization Across Datasets} 
The next measure of successful model performance was presented when it was applied to datasets other than that on which it was trained. Hence, training was performed with each dataset and subsequently evaluated each of these upon all other datasets. This cross evaluation was implemented on the baseline settings and for each of the best training coefficients acquired during coefficient ablation, Figs. \ref{fig:test_f1_best}, \ref{fig:test_sa_best}. The complete ablation study results for the development set and cross evaluation development set results can be found in Tables \ref{tab_cross_eval_test_new}, \ref{tab_cross_eval_dev_new}.  Generalization for the F1 metric is clearly seen in the visual area on the radar chart. \textit{P}IFL-OSCAR displayed the best generalization, followed by magpie{\_}random and IFL-OSCAR-A. 
As seen in Figs \ref{fig:test_f1_best}, \ref{fig:test_sa_best} IFL-C4-A, IFL-OSCAR-A, and SemEval2022, the generalization of models trained on these datasets improved for sequence accuracy. 

\paragraph{Span-Aware Evaluation}
To accompany this qualitative empirical analysis, we used \textit{Refined Span Metrics} (Section~\ref{sec:refined_span_metrics}), the {$\mu$}, {$R$}, and {$GM$} of sequence accuracy and $F1$ to quantitatively show the best generalizing training scheme for the baseline slot loss and the combined SCL-negative best coefficients (Tables~\ref{tab:mu_R_summary},~\ref{tab:scl_neg_cross_manual_mu_R}). These empirical measurements confirm the visualization of \textit{P}IFL-OSCAR generalizing best with RoBERTa over BERT for {$\mu_{GM}$} of SA and F1. These results remain for the best coefficient settings, though at 0.0 the \textit{P}IFL-OSCAR results vary without explanation, shifting the highest score of {$\mu_{GM}$} to BERT.  
\begin{table}[!htbp]
\centering
\small
\resizebox{\linewidth}{!}{
    \begin{tabular}{lcccccccc}
    \toprule
    Model & $\mu_{\mathrm{SA}}$ & $\mu_{\mathrm{F1}}$ & $\mu_{\mathrm{GM}}$ & Hardest ds. & $R_{\mathrm{SA}}$ & $R_{\mathrm{F1}}$ & $R_{\mathrm{GM}}$ \\
    \midrule
    IFL-C4-A-BERT & 0.3762 & 0.3112 & 0.3421 & FLUTE & 0.0968 & 0.0229 & 0.0471 \\
    IFL-C4-A-RoBERTa & 0.3513 & 0.2940 & 0.3214 & SEMEVAL2022 & 0.0862 & 0.0000 & 0.0000 \\
    FLUTE-BERT & 0.1910 & 0.1960 & 0.1935 & SEMEVAL2022 & 0.0906 & 0.0206 & 0.0432 \\
    FLUTE-RoBERTa & 0.1813 & 0.1358 & 0.1569 & PIFL-OSCAR & 0.0508 & 0.0249 & 0.0356 \\
    MAGPIE (random)-BERT & 0.4575 & 0.4558 & 0.4566 & SEMEVAL2022 & 0.1446 & 0.0714 & 0.1016 \\
    MAGPIE (random)-RoBERTa & 0.4459 & 0.4607 & 0.4532 & SEMEVAL2022 & 0.1366 & 0.0689 & 0.0970 \\
    PIFL-OSCAR-BERT &\textbf{ 0.5884} &\textbf{ 0.6045} & {0.5964} & FLUTE & 0.1339 & 0.1256 & 0.1297 \\
    PIFL-OSCAR-RoBERTa & \textbf{0.5883} & {0.6094} & {0.5988} & FLUTE & 0.1313 & 0.1367 & 0.1340 \\
    SEMEVAL-BERT & 0.1796 & 0.0524 & 0.0970 & IFL-C4-A & 0.0101 & 0.0000 & 0.0000 \\
    SEMEVAL-RoBERTa & 0.1907 & 0.0659 & 0.1121 & IFL-C4-A & 0.0092 & 0.0000 & 0.0000 \\
    IFL-OSCAR-A-BERT & 0.3774 & 0.3528 & 0.3649 & FLUTE & 0.1141 & 0.0673 & 0.0876 \\
    IFL-OSCAR-A-RoBERTa & 0.3791 & 0.3568 & 0.3678 & FLUTE & 0.0981 & 0.0296 & 0.0539 \\
    \bottomrule
    \end{tabular}
}
\caption{Summary statistics computed on \textbf{test} splits only. $\mu$ is the mean across the six evaluation datasets (rows). The robustness factor $R$ is the model's performance on the most difficult dataset, as the minimum geometric mean $\sqrt{\mathrm{SA}\cdot\mathrm{F1}}$ across datasets; $R_{\mathrm{SA}}$ and $R_{\mathrm{F1}}$ are the corresponding minima over datasets, and $R_{\mathrm{GM}}=\sqrt{R_{\mathrm{SA}}\cdot R_{\mathrm{F1}}}$.\label{tab:mu_R_summary}}
\end{table}

\begin{table}[!htbp]
\centering
\small
\resizebox{\linewidth}{!}{
    \begin{tabular}{lcccccccc}
    \toprule
    Model & $\mu_{\mathrm{SA}}$ & $\mu_{\mathrm{F1}}$ & $\mu_{\mathrm{GM}}$ & Hardest ds. & $R_{\mathrm{SA}}$ & $R_{\mathrm{F1}}$ & $R_{\mathrm{GM}}$ \\
    \midrule
    IFL-C4-A-BERT 1.0 & 0.3749 & 0.3104 & 0.3202 & FLUTE & 0.0955 & 0.0285 & 0.0522 \\
    IFL-C4-A-RoBERTa 1.0 & 0.3595 & 0.2993 & 0.2916 & SEMEVAL2022 & 0.0862 & 0.0000 & 0.0000 \\
    FLUTE-BERT 0.1 & 0.2253 & 0.2509 & 0.2372 & SEMEVAL2022 & 0.0381 & 0.0258 & 0.0314 \\
    FLUTE-RoBERTa 0.1 & 0.2320 & 0.2257 & 0.2263 & SEMEVAL2022 & 0.0879 & 0.0317 & 0.0528 \\
    MAGPIE (random)-BERT 0.1 & 0.4677 & 0.4811 & 0.4676 & FLUTE & 0.1472 & 0.0783 & 0.1074 \\
    MAGPIE (random)-RoBERTa 0.5 & 0.4670 & 0.4805 & 0.4674 & SEMEVAL2022 & 0.1525 & 0.0363 & 0.0744 \\
    PIFL-OSCAR-BERT 0.0 & \textbf{0.6006} & {0.6160} & {0.6046} & FLUTE & 0.1286 & 0.1280 & 0.1283 \\
    PIFL-OSCAR-RoBERTa 0.0 & \textbf{0.5875} & {0.6113} & {0.5967} & FLUTE & 0.1260 & 0.1315 & 0.1287 \\
    SEMEVAL-BERT 1.0 & 0.1850 & 0.0532 & 0.0671 & IFL-C4-A & 0.0097 & 0.0000 & 0.0000 \\
    SEMEVAL-RoBERTa 0.5 & 0.1936 & 0.0668 & 0.0802 & IFL-C4-A & 0.0099 & 0.0000 & 0.0000 \\
    IFL-OSCAR-A-BERT 0.1 & 0.3809 & 0.3593 & 0.3606 & FLUTE & 0.1141 & 0.0707 & 0.0898 \\
    IFL-OSCAR-A-RoBERTa 0.5 & 0.3816 & 0.3494 & 0.3454 & FLUTE & 0.0942 & 0.0296 & 0.0528 \\
    \bottomrule
    \end{tabular}
}
\caption{SCL-neg (test only). $\mu$ is the mean across evaluation datasets. $\mu_{\mathrm{GM}}$ is the mean of $\sqrt{\mathrm{SA}\cdot\mathrm{F1}}$ across datasets. The hardest dataset is the dataset attaining the minimum per-dataset GM for the model. $R_{\mathrm{SA}}$ and $R_{\mathrm{F1}}$ are the minima across datasets (test), and $R_{\mathrm{GM}}=\sqrt{R_{\mathrm{SA}}\cdot R_{\mathrm{F1}}}$.}
\label{tab:scl_neg_cross_manual_mu_R}
\end{table}

\begin{figure}[!htbp]
    \centering
    \includegraphics[width=1\linewidth]{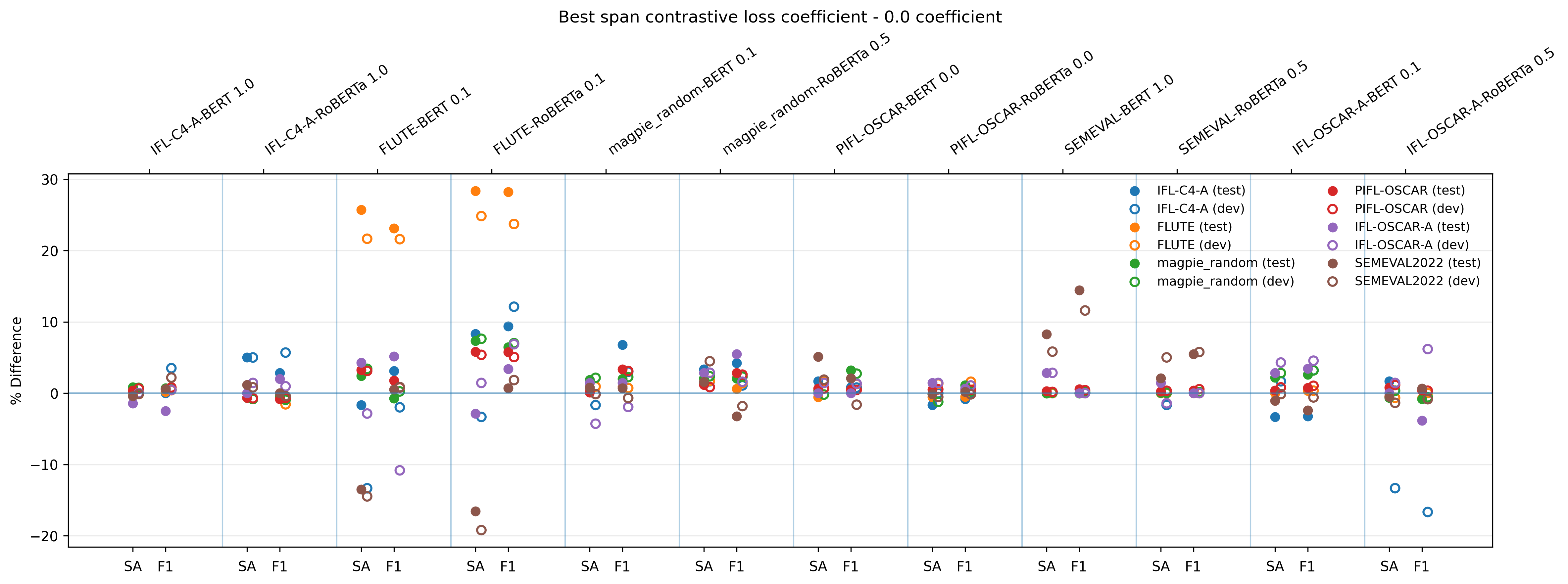}
    \caption{Difference between the best coefficient of span contrastive loss and baseline results.}
    \label{fig:best-base-SA-F1}
\end{figure}

\paragraph{Impact of Span Contrastive Loss}
To confirm the model training from baseline to best coefficient, Fig. \ref{fig:best-base-SA-F1} shows the improvements, though evaluation for SemEval shows minimal increase. To confirm the stability of the training on the development set, the performance difference between the test and dev set results in Figure ($m_{test} - m_{dev}\, \textrm{where}\, m \in \{F1, SA\} $  base) can be seen. Most values are stable for sequence accuracy (SA) and F1 with one significant outlier: IFL-C4-A. The results show low performance with SCL, possibly due to the lack of diversity in the dataset or the difference between the two sets, or the small size of the dataset. SemEval2022 and FLUTE performance and evaluation can also be attributed to this.

\paragraph{Latent Representation Analysis}
From the figures showing the stability of the models in test-dev results, best coefficient - baseline, and the tables showing the overall performance and that over existing SOTA,  significant improvements were observed. While these metrics show satisfactory empirical evaluations, further examination is required to fully grasp the variation in semantic space. For this, PCA and tSNE dimension reduction algorithms were employed to help visualize this shift. PCA retains the global representations as a linear decomposition process, while tSNE helps visualize the local representations by probabilistic compression \cite{PCAcai2022theoreticalfoundationstsnevisualizing}. Both show separation between models and training paradigms. The models trained with SCL show more logarithmic diffusion in PCA, indicating refinement in the semantic representation between slightly varied spans, rather than the uniform, generalized cluster we tend to see with the baseline models trained only on slot loss. Both baseline and highest performing coefficient for BERT and RoBERTa are shown in each figure. Figures for the CLS token, span and word embeddings were extracted for analysis of training effect. Greater variability and separation occurs for the span embeddings, which accompanies more linearity in the vector space. Span embeddings constitute all spans with contiguous identical labels, while word embeddings include all words with any B, I, or O label. The span embeddings were also label agnostic, so the contrastive training effect could be more clearly seen without bias toward correctly predicted sequences.

\begin{figure}[!htbp]
\centering

\begin{subfigure}{0.48\textwidth}
    \centering
    \includegraphics[width=\linewidth]{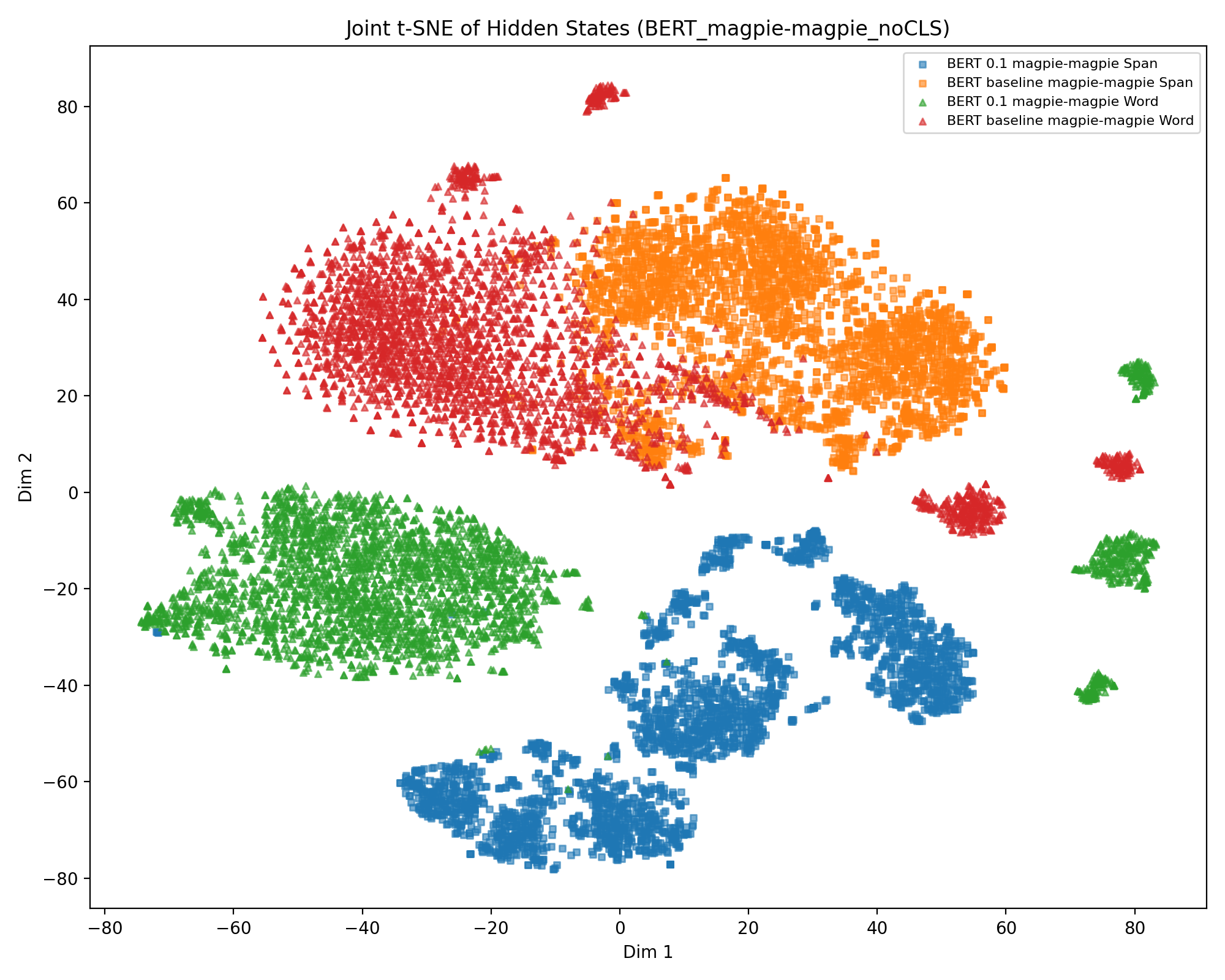}
    \caption{Word and span embedding t-SNE results for MAGPIE (random) trained BERT evaluated on the MAGPIE (random) dataset.}
    \label{fig:bert-magpie-magpie-tsne}
\end{subfigure}
\hfill
\begin{subfigure}{0.48\textwidth}
    \centering
    \includegraphics[width=\linewidth]{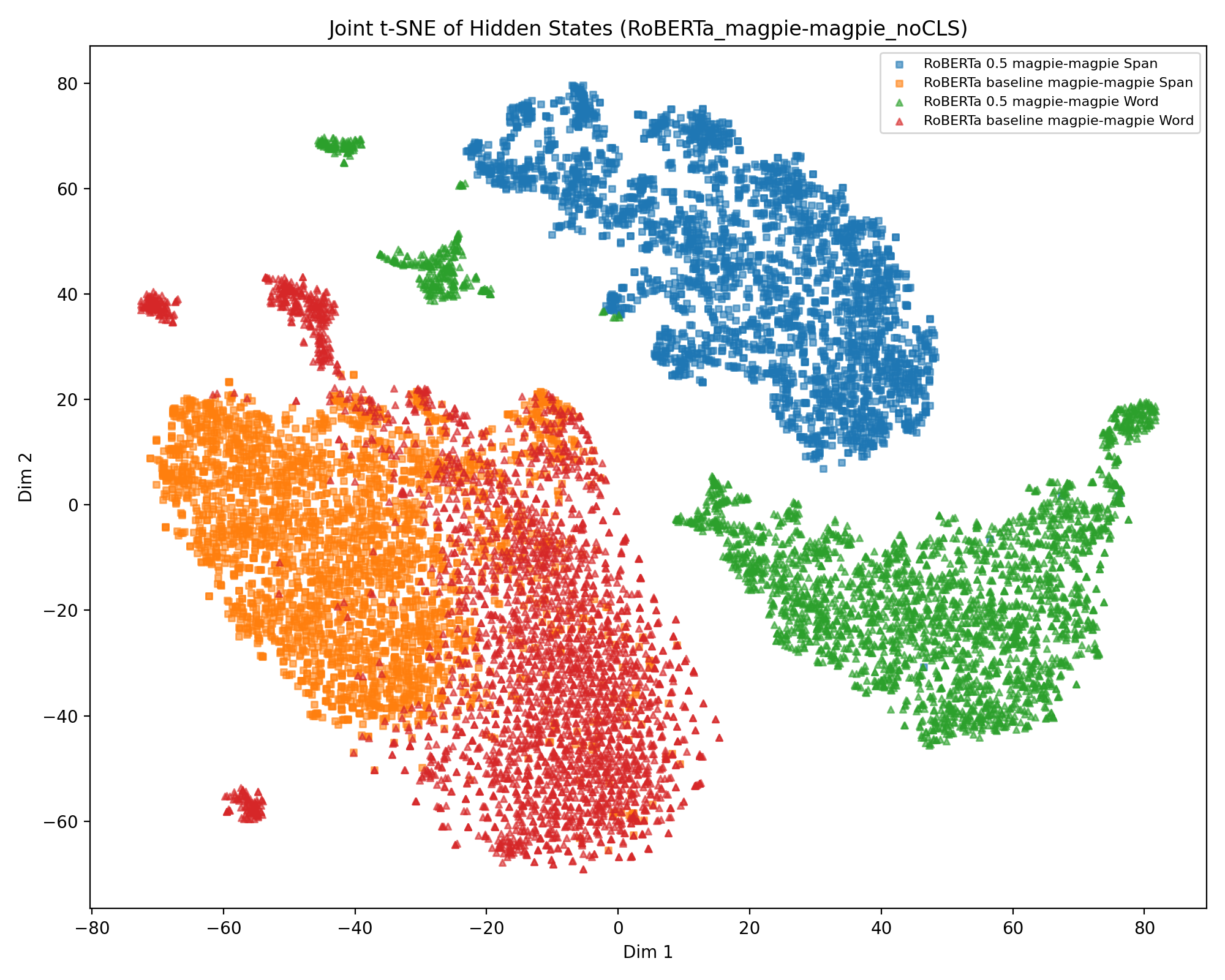}
    \caption{Word and span embedding t-SNE results for MAGPIE (random) trained RoBERTa evaluated on the MAGPIE (random) dataset.}
    \label{fig:roberta-magpie-magpie-tsne}
\end{subfigure}

\caption{t-SNE visualization of word and span embeddings for MAGPIE (random) trained models evaluated on the MAGPIE (random) dataset.}
\label{fig:magpie-magpie-tsne}

\end{figure}
In order to show this more clearly, t-SNE results were included for magpie-magpie for word and span hidden states comparing baseline and best coefficient. BERT and RoBERTa were also compared with PCA. Figures \ref{fig:bert-magpie-magpie-tsne} and \ref{fig:roberta-magpie-magpie-tsne} show their inherent differences and ability to be pushed by contrastive objectives over word embedding space and span embedding space. While t-SNE analysis tends to favor local representations,  overlaps in the baseline compared to clearer separations for the SCL-neg settings indicated an effective objective.

\begin{figure}[!htbp]
\centering

\begin{subfigure}{0.45\textwidth}
    \centering
    \includegraphics[width=\linewidth]{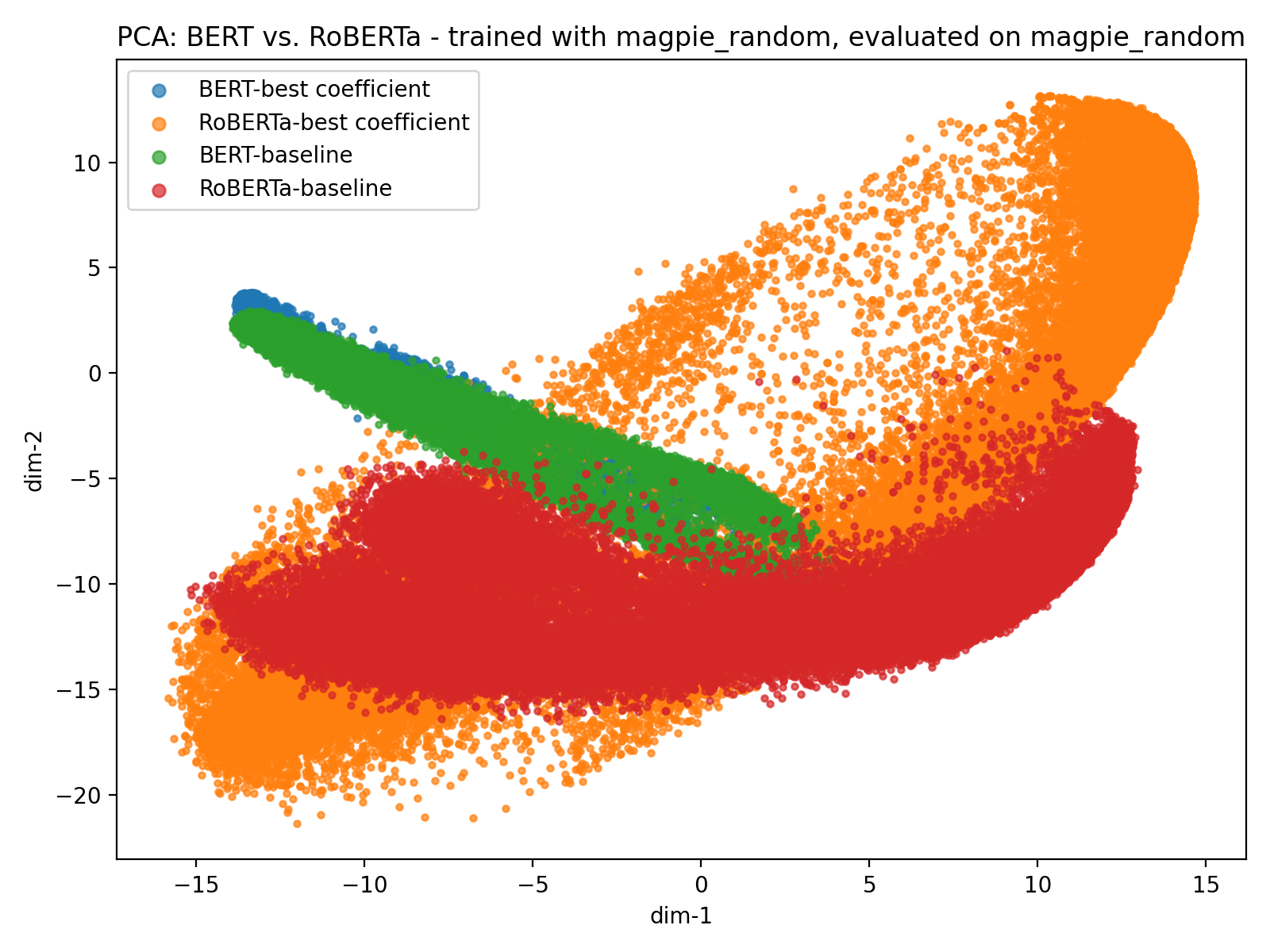}
    \caption{Word embedding PCA results for MAGPIE (random) trained BERT and RoBERTa evaluated on MAGPIE (random) dataset.}
    \label{fig:word_magpie-magpie-pca}
\end{subfigure}
\hfill
\begin{subfigure}{0.45\textwidth}
    \centering
    \includegraphics[width=\linewidth]{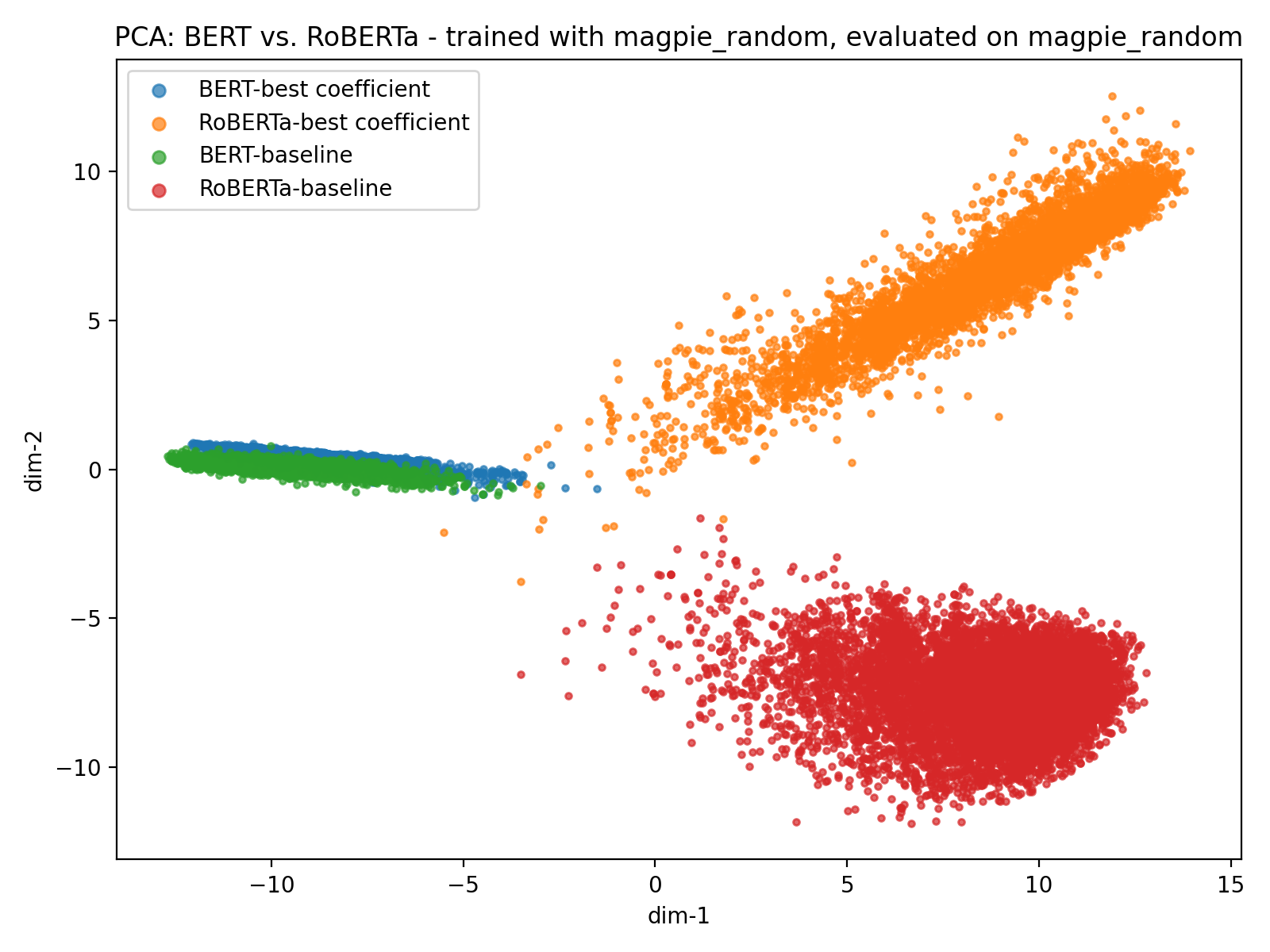}
    \caption{Span embedding PCA results for MAGPIE (random) trained BERT and RoBERTa evaluated on MAGPIE (random) dataset.}
    \label{fig:span-magpie-magpie-pca}
\end{subfigure}

\caption{PCA visualization of embeddings for MAGPIE (random) trained models evaluated on the MAGPIE (random) dataset.}

\end{figure}
Notably, the SemEval dataset contains 50 percent Portuguese examples. While this certainly affects the cross training and evaluation with other models, it can also be seen clearly in the tSNE decomposition figure; two distinct regions appear, especially for BERT-baseline. The PCA decomposition does not capture this nuance at all, lending more relevance to comparative results that found better cluster boundaries in tSNE \cite{TSNEandPCAPareek}. 
Baseline RoBERTa and the models trained on SemEval with SCL-neg, show slightly less cluster coherence in tSNE, perhaps skewing the lines between languages. This cross-lingual span similarity was an untrained outcome that could be explored more deeply in further research.

In each of the PCA adjusted embedding visuals as well as the improvement statistics, we see BERT baseline results nearer to the SCL-neg results than RoBERTa's, Figs. \ref{fig:word_magpie-magpie-pca} and \ref{fig:span-magpie-magpie-pca} . RoBERTa trained models can be seen with greater separation in the embedding space and in the quantified comparative results. The models differ in many substantial ways. The two differences that \textit{should} allow BERT models to be influenced by fine-tuning more greatly than RoBERTa: A larger amount of pretraining data and longer pretraining time.

\subsection{Error Analysis}
Deeper examination of output errors shows variance between BERT and RoBERTa architectures for each baseline model and those trained with SCL-neg and self-evaluated on their respective datasets. The following tables lists a selection of fine-grained errors in two main categories: over prediction (the model predicts an idiomatic expression where none occurs), or under prediction (an idiomatic expression or partial sequence is not labeled). These errors disproportionately affected the sequence accuracy of the evaluation compared to the precision, as a partial match registers for precision, though not an exact span match for sequence accuracy. Examples are also shown to highlight improvements as a result of training with SCL-neg. The base models vary in the typical error type between BERT slot only, RoBERTa slot only, BERT with SCL-neg and RoBERTa with SCL-neg. Pretrained RoBERTa was intended to improve BERT by removing the next sentence prediction objective, increase training time, and increase training data. The tokenizer and tokenized results also vary between the two models. To accommodate for this, our models labeled the first token, but ignored the remaining tokens in a given word, keeping the labels consistent by word rather than by token. The following error examples revealed this explicitly. 
The models trained and evaluated on their respective datasets performed highest, hence the tables below represent a selection for each.
\begin{enumerate}
    \item Over prediction - Errors regarding over prediction involve overfitting. The MWE is labeled correctly though additional words are labeled.
    \item Under prediction - This type of error concerns the model identifying only a part of the target MWE.
    \item SCL-neg improvement - This set of examples shows errors existing for baseline settings, but have been resolved by the application of SCL-neg. 
\end{enumerate}

\begin{table*}[htbp]
\centering
\caption{Grouped examples by dataset and instance ID: sentence, gold labels, and predictions from BERT and RoBERTa.}
\label{tab:grouped-by-dataset-id-2}
\resizebox{\columnwidth}{!}{%
\begin{tabular}{p{0.22\textwidth} p{0.9\textwidth}}
\toprule
\multicolumn{2}{l}{\textbf{MAGPIE (random) - Over prediction}}\\
\midrule
\textit{Sentence} &
i suppose you get used to it after a while , sleeping in the day . \\
\textit{Gold Labels - (none)} &
i suppose you get used to it after a while , sleeping in the day . \\
\textit{BERT baseline - incorrect}&
I suppose you get used to it after a while , B-idiom[sleeping] I-idiom[in] I-idiom[the] I-idiom[day] . \\
\textit{RoBERTa - incorrect}&
I suppose you get used to it after a while , B-idiom[sleeping] I-idiom[in] I-idiom[the] I-idiom[day] . \\
\textit{BERT SCL-neg - incorrect}&
I suppose you get used to it after a while , B-idiom[sleeping] I-idiom[in] I-idiom[the] I-idiom[day] . \\
\textit{RoBERTa SCL-neg - incorrect}&
I suppose you get used to it after a while , B-idiom[sleeping] I-idiom[in] I-idiom[the] I-idiom[day] . \\
\midrule

\multicolumn{2}{l}{\textbf{MAGPIE (random) Under prediction}}\\
\midrule
\textit{Sentence} &
No time like the present for getting things in motion. \\
\textit{Gold Labels} &
B-idiom[No] I-idiom[time] I-idiom[like] I-idiom[the] I-idiom[present] for getting things in motion . \\
\textit{BERT - correct} &
B-idiom[No] I-idiom[time] I-idiom[like] I-idiom[the] I-idiom[present] for getting things in motion . \\
\textit{RoBERTa - incorrect} &
B-idiom[No] I-idiom[time] like the I-idiom[present] for getting things in motion . \\
\textit{BERT SCL-neg - correct} &
B-idiom[No] I-idiom[time] I-idiom[like] I-idiom[the] I-idiom[present] for getting things in motion . \\
\textit{RoBERTa SCL-neg -correct} &
B-idiom[No] I-idiom[time] I-idiom[like] I-idiom[the] I-idiom[present] for getting things in motion .\\\midrule

\multicolumn{2}{l}{\textbf{PIFL-OSCAR - Over prediction}}\\
\midrule
\textit{Sentence} &
a fellow that blows hot and cold in the same breath cannot be friends with me! \\
\textit{Gold Labels} &
a fellow that blows hot and cold B-idiom[in] I-idiom[the] I-idiom[same] I-idiom[breath] cannot be friends with me! \\
\textit{BERT - incorrect}&
a fellow that blows hot I-idiom[and] O[cold] B-idiom[in] I-idiom[the] I-idiom[same] I-idiom[breath] cannot be friends with me! \\
\textit{RoBERTa - correct}&
a fellow that blows hot and cold B-idiom[in] I-idiom[the] I-idiom[same] I-idiom[breath] cannot be friends with me! \\
\textit{BERT SCL-neg - correct} &
a fellow that blows hot and cold B-idiom[in] I-idiom[the] I-idiom[same] I-idiom[breath] cannot be friends with me! \\
\textit{RoBERTa SCL-neg -correct} &
a fellow that blows hot and cold B-idiom[in] I-idiom[the] I-idiom[same] I-idiom[breath] cannot be friends with me!\\
\midrule

\multicolumn{2}{l}{\textbf{PIFL-OSCAR - Over prediction}}\\
\midrule
\textit{Sentence} &
then we start from scratch and design a complete new algorithm , called present , where we could build upon the results of the first step . \\
\textit{Gold Labels} &
As one character notes , this is a school that doesn ' t just B-idiom[think] I-idiom[outside] I-idiom[the] I-idiom[box] - - its leaders " think outside the dictionary . " \\
\textit{BERT - incorrect}&
As one character notes , this is a school that doesn ' t just B-idiom[think] I-idiom[outside] I-idiom[the] I-idiom[box] - - its leaders " B-idiom[think] outside the dictionary . " \\
\textit{RoBERTa - incorrect}&
As one character notes , this is a school that doesn ' t just B-idiom[think] I-idiom[outside] I-idiom[the] I-idiom[box] - - its leaders " B-idiom[think] I-idiom[outside] I-idiom[the] I-idiom[dictionary] . " \\
\textit{BERT SCL-neg - correct/improvement}&
As one character notes , this is a school that doesn ' t just B-idiom[think] I-idiom[outside] I-idiom[the] I-idiom[box] - - its leaders " B-idiom[think] outside I-idiom[the] dictionary . " \\
\textit{RoBERTa SCL-neg -correct/improvement}&
As one character notes , this is a school that doesn ' t just B-idiom[think] I-idiom[outside] I-idiom[the] I-idiom[box] - - its leaders " B-idiom[think] I-idiom[outside] I-idiom[the] I-idiom[dictionary] . "\\
\midrule

\bottomrule

\end{tabular}
}
\end{table*}

\paragraph{Lingering Mistakes}
The final error example represented a typical and enduring challenge for Idiom- and Figurative Language-aware systems. "Think outside the dictionary" is clearly an adaptation of the gold labeled, "think outside the box." While this is easily identifiable and interpretable by a speaker of English, the system considered it an error due to the gold labels. Only RoBERTa effectively defeats this in baseline and SCL-neg settings. BERT SCL-neg improved the label classification accuracy with the exception of the label for the word "dictionary".

\section{Conclusion}
This work resolved questions regarding span modeling with an emphasis on span contrastive loss with hard negatives, coefficient ablation, and cross evaluation. The geometric mean of sequence accuracy and F1, created a new span-aware composite metric that can be adopted for future idiomatic or figurative language research. 

The span contrastive loss addition to the slot loss architecture has proven to be an effective method of increasing the performance of BERT and RoBERTa, models fine-tuned with idiom and figurative language datasets using a supervised learning architecture featuring emphasized hard negatives. Through this combination, sequence tagging and generalization on new and existing datasets was exceptional. Sequence accuracy was further improved to SOTA on MAGPIE-Idioms \textit{random} using slot and span contrastive loss while representing a span as the mean of its constituent embeddings.
Utilizing these objectives also helped establish new benchmarks for \textit{P}IFL-OSCAR, IFL-OSCAR-A, and IFL-C4-A. While these results are desirable, more improvements and efficiencies could be made by utilizing the power and generalizability of larger language models. The FLUTE and SemEval datasets adapted for our purposes were not labeled by their authors for our training scheme, and unsurprisingly resulted in our models' worst performance on any training setting, as shown in the metric ${R}$.


\appendix

\begin{landscape}
\begingroup
\setlength{\tabcolsep}{3.5pt}
\renewcommand{\arraystretch}{1.0}
\small

\begin{center}
\captionof{table}{Cross evaluation results for test set from best coefficient on BERT-based models. For each column, each metric is shown. (Sequence Accuracy/F1/Precision/Recall)}

\label{tab_cross_eval_test_new}
\resizebox{0.98\linewidth}{!}{%
\begin{tabular}{|l|c|c|c|c|c|c|}
\hline
\textbf{Dataset} & \textbf{IFL-C4-A} & \textbf{FLUTE} & \textbf{MAGPIE (random)} & \textbf{PIFL-OSCAR} & \textbf{IFL-OSCAR-A} & \textbf{SemEval-2022} \\
\hline
IFL-C4-A-BERT (1.0)        & 95/96.91/95.92/97.92 & 9.55/3.3/12.38/1.9 & 14.85/21.14/31.75/15.85 & 25.12/37.02/55.51/27.77 & 40/25/33.33/20 & 40.42/2.85/5.71/1.9 \\
\hline
IFL-C4-A-RoBERTa (1.0)     & 91.67/93.75/93.75/93.75 & 8.62/1.56/6.98/0.88 & 12.81/20.34/40.4/13.59 & 24.8/37.24/61.34/26.73 & 38.57/26.67/40/20 & 39.24/0/0/0 \\
\hline
FLUTE-BERT (0.1)           & 26.67/31.67/26.39/39.58 & 61.41/61.81/59.86/63.89 & 16.67/19.95/17.82/22.66 & 12.33/17.49/15.53/20.02 & 14.29/17.02/13.19/24 & 3.81/2.58/1.65/5.94 \\
\hline
FLUTE-RoBERTa (0.1)        & 25/24.18/25.58/22.92 & 54.91/55.54/54.67/56.43 & 16.78/22.08/23.82/20.59 & 10.87/15.7/16.24/15.19 & 22.86/14.75/12.5/18 & 8.79/3.17/1.98/7.86 \\
\hline
MAGPIE (random)-BERT (0.1)   & 40/52.75/55.81/50 & 14.72/15.02/17.01/13.45 & 94.81/95.15/94.44/95.87 & 55.05/62.55/70.15/56.44 & 47.14/55.36/50/62 & 28.87/7.83/8.7/7.13 \\
\hline
MAGPIE (random)-RoBERTa (0.5)& 45/56.25/56.25/56.25 & 15.25/14.23/15.7/13.01 & 94.65/94.93/94.5/95.37 & 55.36/62.13/68.95/56.53 & 50/57.14/51.61/64 & 19.95/3.63/3.29/4.05 \\
\hline
PIFL-OSCAR-BERT (0.0)      & 80/88.89/80/100 & 12.86/12.8/14.58/11.4 & 65.94/71.29/73.32/69.37 & 98.7/99.06/98.78/99.35 & 67.14/80.65/67.57/100 & 35.7/16.9/20.27/14.49 \\
\hline
PIFL-OSCAR-RoBERTa (0.0)   & 78.33/88.07/78.69/100 & 12.6/13.15/14.78/11.84 & 65.92/70.87/73.56/68.37 & 98.22/98.73/98.46/99 & 68.57/81.3/68.49/100 & 28.87/14.67/14.35/15 \\
\hline
SEMEVAL-BERT (1.0)         & 20/0/0/0 & 9.15/0/0/0 & 0.97/0.97/19.13/0.5 & 2.2/3.69/37.09/1.94 & 30/0/0/0 & 56.96/41.73/64.36/30.88 \\
\hline
SEMEVAL-RoBERTa (0.5)      & 20/0/0/0 & 9.15/0/0/0 & 0.99/1.04/13.56/0.54 & 2.23/3.44/28.27/1.83 & 28.57/0/0/0 & 57.35/41.05/58.33/31.67 \\
\hline
IFL-OSCAR-A-BERT (0.1)     & 46.67/46.15/60/37.5 & 11.41/7.07/18.18/4.39 & 23.12/30.88/40.12/25.1 & 29.5/41.47/54/33.66 & 78.57/79.61/77.36/82 & 38.19/8/13.41/5.7 \\
\hline
IFL-OSCAR-A-RoBERTa (0.5)  & 56.67/60/75/50 & 9.42/4.16/9.94/2.63 & 21.23/30.13/45.02/22.64 & 28.86/41.22/57.93/32 & 72.86/71.15/68.52/74 & 39.37/3.6/7.35/2.38 \\
\hline
\end{tabular}
}
\end{center}

\vspace{6mm}

\begin{center}
\captionof{table}{Cross evaluation results for development set from best coefficient on BERT-based models. For each column, each metric is shown. (Sequence Accuracy/F1/Precision/Recall)}

\label{tab_cross_eval_dev_new}
\resizebox{0.98\linewidth}{!}{%
\begin{tabular}{|l|c|c|c|c|c|c|}
\hline
\textbf{Dataset} & \textbf{IFL-C4-A} & \textbf{FLUTE} & \textbf{MAGPIE (random)} & \textbf{PIFL-OSCAR} & \textbf{IFL-OSCAR-A} & \textbf{SemEval-2022} \\
\hline
IFL-C4-A-BERT (1.0)        & 88.33/92.63/86.27/100 & 11.42/3.87/15.15/2.22 & 15.32/21.68/31.62/16.49 & 25.02/36.83/53.82/28 & 61.43/54.12/57.5/51.11 & 45.74/2.18/5.21/1.38 \\
\hline
IFL-C4-A-RoBERTa (1.0)     & 86.67/91.11/89.13/93.18 & 11.16/3.19/15.58/1.78 & 13.8/21.46/41.58/14.47 & 24.19/36.73/60.32/26.41 & 51.43/35.82/54.55/26.67 & 44.38/0/0/0 \\
\hline
FLUTE-BERT (0.1)           & 13.33/21.31/16.67/29.55 & 60.03/60.57/59.49/61.69 & 17.07/21.03/19.07/23.43 & 12.5/16.96/15.09/19.36 & 18.57/12.5/9.09/20 & 4.6/2.24/1.37/6.08 \\
\hline
FLUTE-RoBERTa (0.1)        & 23.33/29.7/26.32/34.09 & 54.58/55.43/54.68/56.21 & 16.98/22.25/23.68/20.98 & 11.08/15.55/16.24/14.92 & 31.43/23.53/21.05/26.67 & 7.98/2.31/1.4/6.63 \\
\hline
MAGPIE (random)-BERT (0.1)   & 35/49.46/46.94/52.27 & 15.67/15.71/18.31/13.76 & 94.68/94.98/94.41/95.56 & 55.33/62.58/69.58/56.85 & 50/63.55/54.84/75.56 & 31.39/5.81/6.13/5.52 \\
\hline
MAGPIE (random)-RoBERTa (0.5)& 43.33/54.55/49.09/61.36 & 15.8/15.15/16.85/13.76 & 94.62/95.01/94.7/95.34 & 55.09/61.42/67.87/56.1 & 51.43/63.55/54.84/75.56 & 23.14/3.53/3.07/4.14 \\
\hline
PIFL-OSCAR-BERT (0.0)      & 70/83.02/70.97/100 & 16.73/16.12/18.15/14.5 & 65.83/71.07/73.52/68.77 & 98.99/99.27/99.06/99.5 & 62.86/77.59/63.38/100 & 35.45/9.97/11/9.12 \\
\hline
PIFL-OSCAR-RoBERTa (0.0)   & 68.33/82.24/69.84/100 & 15.54/15.45/17.15/14.05 & 65.54/70.74/73.35/68.3 & 98.41/98.91/98.58/99.25 & 62.86/77.59/63.38/100 & 28.28/8.22/7.67/8.84 \\
\hline
SEMEVAL-BERT (1.0)         & 26.67/0/0/0 & 10.23/0/0/0 & 0.96/1.1/20/0.56 & 1.92/3.3/38.04/1.72 & 37.14/0/0/0 & 59.54/38.36/58.86/28.45 \\
\hline
SEMEVAL-RoBERTa (0.5)      & 26.67/0/0/0 & 10.09/0/0/0 & 0.81/0.83/11.31/0.43 & 2.12/3.5/29.21/1.86 & 34.29/0/0/0 & 60.76/39.43/56.12/30.39 \\
\hline
IFL-OSCAR-A-BERT (0.1)     & 53.33/47.37/56.25/40.91 & 11.55/5.92/14.79/3.7 & 24.5/31.9/40.32/26.39 & 30.48/42.64/54.76/34.92 & 84.29/81.63/75.47/88.89 & 44.93/4.95/9.76/3.31 \\
\hline
IFL-OSCAR-A-RoBERTa (0.5)  & 50/50/55.56/45.45 & 11.69/6.9/15.54/4.44 & 22.35/31.18/45.39/23.75 & 29.36/41.25/57.21/32.25 & 74.29/75/70.59/80 & 42.08/0.41/0.77/0.28 \\
\hline
\end{tabular}
}
\end{center}

\clearpage
\endgroup
\end{landscape}

\bibliography{ref}
\end{document}